\pdfoutput=1

\documentclass[11pt]{article}

\usepackage[final]{acl}

\usepackage{times}
\usepackage{arydshln}
\usepackage{latexsym}
\usepackage{amssymb}
\usepackage{enumitem}

\usepackage[T1]{fontenc}

\usepackage[utf8]{inputenc}

\usepackage{microtype}

\usepackage{inconsolata}

\usepackage{multirow}
\usepackage{adjustbox}
\usepackage{xcolor}
\usepackage{booktabs}

\usepackage{xspace}
\newcommand{\benchmark}{LaMP\xspace}

\title{\benchmark: When Large Language Models Meet Personalization}

\author{Alireza Salemi\textsuperscript{1}, Sheshera Mysore\textsuperscript{1}, Michael Bendersky\textsuperscript{2}, Hamed Zamani\textsuperscript{1} \\
  \textsuperscript{1}University of Massachusetts Amherst \\
  \textsuperscript{2}Google Research \\
  \texttt{\{asalemi,smysore,zamani\}@cs.umass.edu} \\ \texttt{bemike@google.com}\\ \\
  The LaMP Benchmark: \url{http://lamp-benchmark.github.io/}
}

\begin{document}
\maketitle

\begin{abstract}
This paper highlights the importance of personalization in large language models and introduces the \benchmark benchmark --- a novel benchmark for training and evaluating language models for producing personalized outputs. \benchmark offers a comprehensive evaluation framework with diverse language tasks and multiple entries for each user profile. It consists of seven personalized tasks, spanning three text classification and four text generation tasks. We additionally propose two retrieval augmentation approaches that retrieve personal items from each user profile for personalizing language model outputs. To this aim, we study various retrieval models, including term matching, semantic matching, and time-aware methods. 
Extensive experiments on \benchmark for zero-shot and fine-tuned language models demonstrate the efficacy of the proposed retrieval augmentation approach and highlight the impact of personalization in various natural language tasks.
\end{abstract}

\section{Introduction}

The recent development of large language models (LLMs) has revolutionized natural language processing (NLP) applications. As the use of LLMs, such as GPT-4 \cite{openai2023gpt4}, in real-world applications evolves, personalization emerges as a key factor in meeting the user's expectations for tailored experiences that align with their unique needs and preferences \cite{huang2022usernlp}. Personalization has been widely studied by various communities, including the information retrieval (IR) and human-computer interaction (HCI) communities, often with applications to search engines and recommender systems \cite{10.1145/2702123.2702503, 10.1145/1462198.1462203, naumov2019deep}. Recent work has also highlighted the impact and concerns associated with personalizing LLMs and tying it to ongoing work on alignment \cite{kirk2023personalisation}. Despite this and the importance of personalization in many real-world problems, developing and evaluating LLMs for producing personalized responses remain relatively understudied. To bridge this gap, this paper underscores the importance of personalization in shaping the future of NLP systems and takes a first step towards developing and evaluating personalization in the context of large language models by introducing the {\benchmark} benchmark\footnote{LaMP stands for \underline{La}nguage \underline{M}odel \underline{P}ersonalization.} --- a comprehensive and diverse benchmarks of personalized text classification and generation tasks.

While many existing well-known NLP benchmarks, such as GLUE \cite{glue}, SuperGLUE \cite{superglue}, KILT \cite{kilt}, and GEM \cite{gem} have led to significant progress in various NLP tasks, they have often taken the dominant NLP approach of ``one-size-fits-all'' to modeling and evaluation, and do not allow the development of models that adapt to the specific needs of end users -- limiting extensive research on personalization in NLP tasks. In contrast, \benchmark offers a comprehensive evaluation framework incorporating diverse language tasks that require personalization. \benchmark consists of three personalized text classification tasks: (1) Personalized Citation Identification (binary classification), (2) Personalized Movie Tagging (categorical classification with 15 tags), and (3) Personalized Product Rating (ordinal classification from 1 to 5-star rating for e-commerce products). Further, \benchmark includes four text generation datasets: (4) Personalized News Headline Generation, (5) Personalized Scholarly Title Generation, (6) Personalized Email Subject Generation, and (7) Personalized Tweet Paraphrasing. For these seven tasks, we explore the two dominant settings in personalization: (a) personalization for new users with a user-based data split and (b) personalization for future interactions of existing users with a time-based data split.
Therefore, \benchmark provides a rich environment for developing personalized NLP models. To foster research in this area, we release the \benchmark benchmark, the data construction, evaluation scripts, and a leaderboard.

For personalizing the language model outputs, a straightforward solution is to incorporate the user profile into a language model prompt. However, user profiles are often large and exceed the length limitations of large language models. Even as such limitations are relaxed with evolving technology, the cost of processing large input sequences is considerable. Therefore, we propose two retrieval augmentation solutions for LLM personalization, in which for each test input, we retrieve items from the user profile to be included in the LLM prompt for personalization. The first approach uses in-prompt augmentation (IPA) for personalization, and the second approach encodes each personal item separately and integrate them later in the decoder using the fusion-in-decoder model of \citet{fid}. We demonstrate that using this approach, the performance of language models improves on all datasets in the \benchmark benchmark. Based on this retrieval augmentation solution, we evaluate different retrievers for personalized prompt construction and establish benchmark results for fine-tuned and zero-shot language models. The empirical findings of our research reveal that the process of fine-tuning a language model utilizing our personalized augmentation technique yields a noteworthy relative average enhancement of 23.5\% across the benchmark. Even in zero-shot settings where an off-the-shelf LLM without fine-tuning (e.g., FlanT5-XXL) is used, utilizing our proposed method results in a relative average improvement of 12.2\% across the tasks. Finally, this paper smooths the path towards developing advanced user-centric NLP systems.

\section{The \benchmark Benchmark}

\paragraph{Problem Formulation.} Generative language models often take an input $x$ and predict the most probable sequence tokens $y$ that follows $x$. Personalizing language models can be defined as conditioning the model's output on a user $u$, represented by a user profile. In \benchmark, we define user profile as the user's historical data, i.e., the past input and personalized outputs produced by or approved by the user, $P_u = \{(x_{u1}, y_{u1}), (x_{u2}, y_{u2}), \cdots, (x_{um_u}, y_{um_u})\}$. Therefore, each data entry in the \benchmark benchmark consists of three components: an input sequence $x$ that serves as the model's input, a target output $y$ that the model is expected to produce, and a profile $P_u$ that encapsulates any auxiliary information that can be used to personalize the model for the user. 

\paragraph{Overview of \benchmark.} Given the above problem formulation, we develop the \benchmark benchmark that aims to assess the efficacy of LLMs in producing personalized outputs $y$, based on inputs $x$ and user-specific information $P_u$. Outputs $y$ of different types result in seven diverse tasks spanning personalized text classification and generation:
\begin{itemize}[noitemsep]
    \item \textbf{Personalized Text Classification}
    \begin{enumerate}
        \item[(1)] Personalized Citation Identification
        \item[(2)] Personalized Movie Tagging
        \item[(3)] Personalized Product Rating
    \end{enumerate}
    \item \textbf{Personalized Text Generation}
    \begin{enumerate}
        \item[(4)] Personalized News Headline Generation
        \item[(5)] Personalized Scholarly Title Generation
        \item[(6)] Personalized Email Subject Generation
        \item[(7)] Personalized Tweet Paraphrasing
    \end{enumerate}
\end{itemize}


\begin{table*}
    \centering
    \begin{adjustbox}{max width=\textwidth}    
        \begin{tabular}{lllccccccc}
        \toprule
            \textbf{Task} & \textbf{Type} & \textbf{Separation} & \textbf{\#Train} & \textbf{\#Dev} & \textbf{\#Test} & \textbf{Input Length} & \textbf{Output Length} & \textbf{Profile Size} & \textbf{\#Classes}\\
            \midrule
            \multirow{2}{*}{Citation Ident.} & \multirow{2}{*}{binary classification} & user & 9682 & 2500 & 2500 & 51.40 $\pm$ 5.72 & \multirow{2}{*}{-} & 90.61 $\pm$ 53.87 & \multirow{2}{*}{2} \\
            & & time & 6542 & 1500 & 1500 & 51.43 $\pm$ 5.70 & & 84.15 $\pm$ 47.54 & \\
            \hline
            \multirow{2}{*}{Movie Tag.} & \multirow{2}{*}{categorical classification} & user & 3820 & 692 & 870 & 92.27 $\pm$ 20.83 & \multirow{2}{*}{-} & 159.29 $\pm$ 330.81 & \multirow{2}{*}{15} \\
            & & time & 5073 & 1410 & 1557 & 92.39 $\pm$ 21.95 & & 86.76 $\pm$ 189.52 & \\
            \hline
            \multirow{2}{*}{Product Rat.} & \multirow{2}{*}{ordinal classification} & user & 20000 & 2500 & 2500 & 145.14 $\pm$ 157.96 & \multirow{2}{*}{-} & 188.10 $\pm$ 129.42 & \multirow{2}{*}{5} \\
            & & time & 20000 & 2500 & 2500 & 128.18 $\pm$ 146.25 & & 185.40 $\pm$ 129.30 & \\
            \hline
            \multirow{2}{*}{News Headline} & \multirow{2}{*}{text generation} & user & 12527 & 1925 & 2376 & 30.53 $\pm$ 12.67 & 9.78 $\pm$ 3.10 & 287.16 $\pm$ 360.62 & \multirow{2}{*}{-} \\
            & & time & 12500 & 1500 & 1800 & 29.97 $\pm$ 12.09 & 10.07 $\pm$ 3.10 & 204.59 $\pm$ 250.75 & \\
            \hline
            \multirow{2}{*}{Scholarly Title} & \multirow{2}{*}{text generation} & user & 9682 & 2500 & 2500 & 152.81 $\pm$ 86.60 & 9.26 $\pm$ 3.13 & 89.61 $\pm$ 53.87 & \multirow{2}{*}{-} \\
            & & time & 14682 & 1500 & 1500 & 162.34 $\pm$ 65.63 & 9.71 $\pm$ 3.21 & 87.88 $\pm$ 53.63 & \\
            \hline
            \multirow{2}{*}{Email Subject} & \multirow{2}{*}{text generation} & user & 4840 & 1353 & 1246 & 436.15 $\pm$ 805.54 & 7.34 $\pm$ 2.83 & 80.72 $\pm$ 51.73 & \multirow{2}{*}{-} \\
            & & time & 4821 & 1250 & 1250 & 454.87 $\pm$ 889.41 & 7.37 $\pm$ 2.78 & 55.67 $\pm$ 36.32 & \\
            \hline
            \multirow{2}{*}{Tweet Para.} & \multirow{2}{*}{text generation} & user & 10437 & 1500 & 1496 & 29.76 $\pm$ 6.94 & 16.93 $\pm$ 5.65 & 17.74 $\pm$ 15.10 & \multirow{2}{*}{-} \\
            & & time & 13437 & 1498 & 1500 & 29.72 $\pm$ 7.01 & 16.96 $\pm$ 5.67 & 15.71 $\pm$ 14.86 & \\
        \bottomrule
        \end{tabular}
    \end{adjustbox}
    \caption{Data statistics of the tasks in the \benchmark benchmark. Each dataset in the \benchmark benchmark has two evaluation settings: (a) user-based data split to test personalization for new users and (b) a time-based data split to test personalization for future interactions of existing users.}
    \label{tab:task-stats}
\end{table*}

\subsection{Tasks Definitions}
Next, we overview of each task used in \benchmark and detail data construction in Appendix \ref{appendix:lamp-data-creation}.

\subsubsection*{\benchmark-1: Personalized Citation Identification}
The citation behavior of researchers is dependent on their interests and is commonly used to evaluate and develop personalized systems for recommending papers \cite{farber2020citrec}. This task recasts citation recommendation as a binary classification task and assesses the ability of a language model to identify user preferences for citations. Specifically, if the user $u$ writes a paper $x$, a language model must determine which of two candidate papers $u$ will cite in $x$ (see Figure \ref{fig:tasks_overview}).

To generate data samples, we leverage the Citation Network Dataset (V14) \cite{Tang:08KDD}, which comprises information on scientific papers, authors, and citations.  
%
For this task, the profile of each user encompasses all the papers they have authored. We retain only the title and abstract of each paper in the user's profile. 

\subsubsection*{\benchmark-2: Personalized Movie Tagging}
Users tagging behavior for media such as movies and books are known to be idiosyncratic and depends on their understanding of the tag and the aspects of the item they focus on. This has motivated a large body of work on personalized tagging \cite{gupta2010socialtags}. We use this task to evaluate the ability of language models to make tag assignments for a movie contingent on the user's historical tagging behavior. Specifically, given a movie description $x$ and a user's historical movie-tag pairs, a language model must predict one of 15 tags for $x$. We obtain tag assignments from the MovieLens dataset \cite{movie-lense}. Additionally, we obtain movie descriptions from MovieDB.\footnote{\url{https://www.themoviedb.org/}} 

\subsubsection*{\benchmark-3: Personalized Product Rating}
Product reviews commonly express a nuanced set of user preferences for a product and which in turn determine their rating for the product. Predicting ratings based on user reviews has been studied extensively in personalized sentiment prediction tasks \cite{mireshghallah2022useridentifier}. While this is commonly treated as a regression task, to use autoregressive language models, we frame it as a multi-class classification task. Specifically, given the user $u$'s historical review and rating pairs and an input review $x$, the model must predict an integer rating from $1-5$. We construct our dataset from a dataset of Amazon reviews \cite{amazon-review}.

\subsubsection*{\benchmark-4: Personalized News Headline Generation}
Authors writing displays distinct stylistic elements influenced by both personal and social factors \cite{zhu2021idiosyncratic}. Journalists authoring headlines are likely to balance between faithfully representing an article, appealing to their readers, and maintaining their own identity. This offers a useful testbed for personalized text generation. Here, we evaluate the ability of a language model to capture the stylistic patterns of an author by requiring it to generate a headline for an input news article, $x$, given a user profile of the authors' historical article-title pairs. To create a dataset, we use a collection of Huffington Post articles \cite{misra2022news, misra2021sculpting}.



\subsubsection*{\benchmark-5: Personalized Scholarly Title Generation}
As with \benchmark-4, the generation of titles for research articles offers a test bed for personalized text generation but varies in text domain. In this task, we require language models to generate titles for an input article $x$, given a user profile of historical article-title pairs for an author. Here, only use article abstracts. We create our dataset from the Citation Network Dataset (V14) \cite{Tang:08KDD} also used for \benchmark-1.

\subsubsection*{\benchmark-6: Personalized Email Subject Generation}
Similar to \benchmark-4 and 5, generating email subjects also provides a valuable test bed for personalized text generation. Email assistance is also known to be a task that significantly benefits from personalization \cite{trajanovski2021microsoftcontext}. Here, we require language models to generate an email subject for an input email message $x$, given historical email-subject pairs authored by a user. For this task, we leverage a private dataset of emails the Avocado Research Email Collection \cite{avocado}. Given its private nature this is unlikely to be contained in pre-training data providing a meaningful challenge for language models. 


\subsubsection*{\benchmark-7: Personalized Tweet Paraphrasing}
Social media posts adhere strongly to various personal stylistic patterns of authors \cite{zhu2021idiosyncratic}. Here, we construct a personalized tweet paraphrasing task and require models to generate a tweet in the style of a user given an input tweet $x$, and a user profile of historical tweets by the user. To construct this task we use data from the Sentiment140 dataset \cite{Sentiment140}. Figure \ref{fig:tasks_overview} provides examples of tasks 1-7 in \benchmark.


\subsection{Data Splits}
To enable evaluation in common personalization settings, \benchmark offers two different data splitting settings: (1) user-based separation and (2) time-based separation. In user-based separation (denoted as LaMP-iU for Task $i$), train/validation/test splits are made by partitioning across users, ensuring that no shared users appear across splits. This strategy measures personalization for new users. 

In time-based separation (denoted as LaMP-iT for Task $i$), train/validation/test splits are made by partitioning user items ordered by time. The most recent user items are chosen to create the input-output pairs, with older items serving as user profiles. Appendix \ref{appendix:lamp-data-creation} contains additional details, and Table \ref{tab:task-stats} reports dataset sizes.

\subsection{Evaluation}

For evaluating classification tasks, we use Accuracy for \benchmark-1 (balanced binary classification), Accuracy/F1 for \benchmark-2 (multi-class classification), and MAE/RMSE for \benchmark-3 (ordinal multi-class classification). Following previous works \cite{paraphrase, headline-generation-2} on text generation, we use Rouge-1/Rouge-L \cite{rouge} as evaluation metrics for the text generation generation tasks (\benchmark-4 to \benchmark-7).

\begin{figure}
    \centering
    \includegraphics[width=\columnwidth]{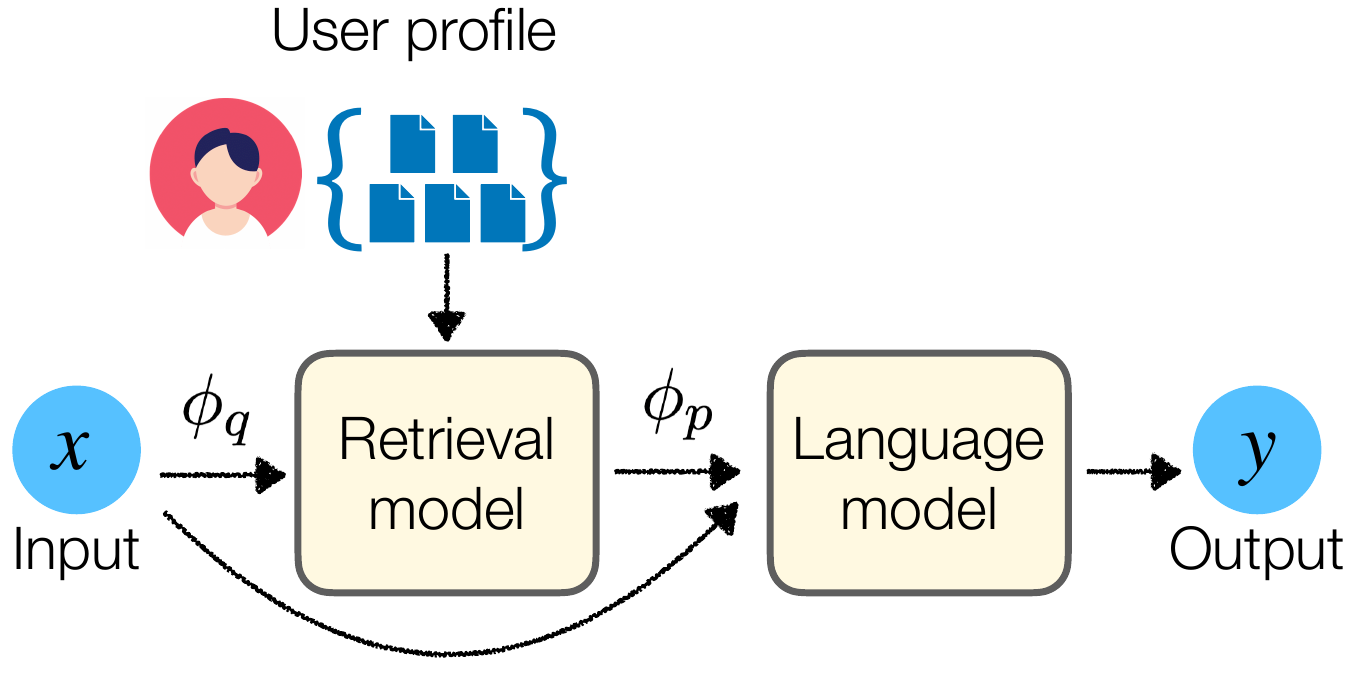}
    \caption{An overview of the retrieval-augmented method for personalizing LLMs. $\phi_q$ and $\phi_p$ represent query and prompt construction functions.}
    \label{fig:overview}
\end{figure}

\section{Retrieval Augmentation for Personalizing LLMs}
\label{sec:profile-method}

To personalize a language model two broad strategies may be explored: (1) fine-tuning the LM for each user and (2) prompting a shared LM with user specific input or context. The former approach necessitates substantial computational resources, especially for fine-tuning larger LLMs. Moreover, accommodating personalized LLMs for each user in industry-scale systems encompassing millions or billions of users necessitates a significant storage and serving capacity. Therefore, we focus on developing strategies for training models personalized via user-specific inputs.

Each task in \benchmark, each user profile consists of a potentially large collection of data points. Given the inherent context length constraint of many LLMs and the cost of processing long sequences, we incorporate a subset of these data points as input prompts. Further, not all entries within a user profile are necessarily relevant to the specific input at hand. To do this, we propose solutions based on retrieval augmentation (See Figure \ref{fig:overview}). This framework selectively extracts pertinent information from the user profile that is relevant to the current unseen test case and generates model predictions conditioned on this information.


Specifically, for a given sample $(x_i, y_i)$ for user $u$, we employ three primary components: (1) a query generation function $\phi_q$ that transforms the input $x_i$ into a query $q$ for retrieving from the user $u$'s profile, (2) a retrieval model $\mathcal{R}(q, P_u, k)$ that accepts a query $q$, a user profile $P_u$ and retrieves $k$ most pertinent entries from the user profile, and (3) a prompt construction function $\phi_p$ that assembles a personalized prompt for the user $u$ based on input $x_i$ and the retrieved entries. For retrieval augmentation, we explore two strategies: \textbf{(1) In-Prompt Augmentation (IPA)} and \textbf{(2) Fusion-in-Decoder (FiD)} \cite{fid}. The input to both approaches constructs inputs, $\bar{x_i}$, using $\mathcal{R}$ to select $k$ items from the user profile $P_u$: 
\begin{equation}
\bar{x_i} = \phi_p(x_i, \mathcal{R}(\phi_q(x_i), P_u, k))
\end{equation}
where we use $(\bar{x_i}, y_i)$ to train or evaluate the language models. With FiD, LLMs receive multiple inputs, each of which is encoded separately within its encoder. These separate encodings are then merged together in the decoder. Here, the inputs $\{\bar{x}_{i1},...,\bar{x}_{ik}\}$ for the encoder are derived as:
\begin{equation}
\bar{x}_{ij} = \phi_p(x_i, d_{ij})
\end{equation}
where $d_{ij}$ is the $j$'th retrieved item using retrieval model from the user profile (i.e, $\mathcal{R}(\phi_q(x_i), P_u, k)$). Note that, IPA and FiD offer different tradeoffs. FiD necessitates training of the language model while IPA may be applied without training. Further, while FiD can only be used with encoder-decoder models, IPA can be used across architectures. However, FiD allows us to incorporate more items from the user profile into the LLM's input.

We explore various choices for the retrieval model $\mathcal{R}$. In our experiments, we study a strong term matching model, BM25 \cite{Robertson1995OkapiBM25}, a state-of-the-art pre-trained dense retrieval model, Contriever \cite{contriever}, a retrieval model that returns most recent profile entries in descending order (i.e., Recency), and a Random document selector from the user profile. The prompt construction function $\phi_p$ \texttt{concat}enates the instruction for each task, the input sequence, and the user profile. The specific prompts are presented in Table \ref{tab:task-prompts}. For the $\phi_q$ function, we use the target input for each task as the query (see Figure \ref{fig:tasks_overview}).

\begin{table*}[!t]
    \centering
    \adjustbox{max width=\textwidth}{
        \begin{tabular}{llccccccc}
        \toprule
         &  & \multicolumn{7}{c}{{FlanT5-base (fine-tuned)}} \\
         \cmidrule(lr){3-9}
        & & \multicolumn{2}{c}{{Non-Personalized}} & \multicolumn{3}{c}{{Untuned profile, $k = 1$}} & \multicolumn{2}{c}{{Tuned profile}} \\
        \cmidrule(lr){3-4} \cmidrule(lr){5-7} \cmidrule(lr){8-9}
         {{Dataset}} & {{Metric}} & No-Retrieval & Random & {Random} & {BM25} & {Contriever} & IPA & FiD($k = 16$) \\\midrule

        \multirow{2}{*}{\shortstack[l]{{\benchmark-1U: Personalized}\\{Citation Identification}}} & \multirow{2}{*}{Accuracy $\uparrow$} & \multirow{2}{*}{0.518} & \multirow{2}{*}{0.539} & \multirow{2}{*}{0.598} & \multirow{2}{*}{0.649} & \multirow{2}{*}{0.688} & \multirow{2}{*}{{0.734}} & \multirow{2}{*}{\textbf{0.754}} \\\\\midrule

        \multirow{2}{*}{\shortstack[l]{{\benchmark-2U: Personalized}\\{Movie Tagging}}} & Accuracy $\uparrow$ & 0.468 & 0.442 & 0.497 & 0.524 & 0.536 & {0.556} & \textbf{0.642} \\
        & F1 $\uparrow$ & 0.435 & 0.403 & 0.459 & 0.480 & 0.506 & {0.519} & \textbf{0.607} \\\midrule

        \multirow{2}{*}{\shortstack[l]{{\benchmark-3U: Personalized}\\{Product Rating}}} & MAE $\downarrow$ & 0.275 & 0.286 & 0.284 & 0.258 & 0.248 & {0.246} & \textbf{0.236} \\
        & RMSE $\downarrow$ & 0.581 & 0.607 & 0.602 & 0.573 & {0.563} & {0.565} & \textbf{0.539} \\\midrule

        \multirow{2}{*}{\shortstack[l]{{\benchmark-4U: Personalized}\\{News Headline Generation}}} & ROUGE-1 $\uparrow$ & 0.153 & 0.159 & 0.162 & 0.167 & 0.173 & \textbf{0.186} & 0.180 \\
        & ROUGE-L $\uparrow$ & 0.140 & 0.147 & 0.148 & 0.153 & 0.159 & \textbf{0.171} & 0.166 \\\midrule

        \multirow{2}{*}{\shortstack[l]{{\benchmark-5U: Personalized}\\{Scholarly Title Generation}}} & ROUGE-1 $\uparrow$ & 0.418 & 0.408 & 0.409 & 0.440 & 0.431 & \textbf{0.450} & 0.431 \\
        & ROUGE-L $\uparrow$ & 0.378 & 0.370 & 0.371 & 0.399 & 0.393 & \textbf{0.409} & 0.392 \\\midrule

        \multirow{2}{*}{\shortstack[l]{{\benchmark-6U: Personalized}\\{Email Subject Generation}}} & ROUGE-1 $\uparrow$ & 0.379 & 0.473 & 0.486 & 0.586 & 0.572 & \textbf{0.587} & 0.567 \\
        & ROUGE-L $\uparrow$ & 0.358 & 0.457 & 0.470 & 0.570 & 0.558 & \textbf{0.575} & 0.555 \\\midrule

        \multirow{2}{*}{\shortstack[l]{{\benchmark-7U: Personalized}\\{Tweet Paraphrasing}}} & ROUGE-1 $\uparrow$ & 0.509 & 0.510 & 0.514 & 0.521 & 0.524 & \textbf{0.528} & 0.517 \\
        & ROUGE-L $\uparrow$ & 0.455 & 0.457 & 0.460 & 0.468 & 0.471 & \textbf{0.475} & 0.464 \\
        \bottomrule
        \end{tabular}
    }
    \caption{The results for a fine-tuned LM on the test set of the user-based setting. The number of retrieved document for personalizing LM is denoted by $k$. Details for tuning the profile on validation sets is in Table \ref{tab:supervised-dev} in Appendix \ref{appendix:dev}.}
    \label{tab:supervised}
\end{table*}

\section{Experiments}

This section describes our experiments, results, and findings on the \benchmark benchmark. 

\subsection{Experimental Setup}

For training FiD and the generative model in IPA we leverage AdamW \cite{adamw} with a learning rate of $5\times10^{-5}$ and a batch size 64 and set 5\% of the total training steps as warmup using a linear scheduler. A weight decay of $10^{-4}$ is incorporated to prevent overfitting. The maximum input and output lengths is set to 512 and 128 tokens, respectively. We train both classification and generation models for 10 and 20 epochs, respectively. A FlanT5-base\footnote{\url{https://huggingface.co/google/flan-t5-base}} \cite{flant5} model is used for all experiments, unless explicitly stated otherwise (in experiments with LLMs, we use FlanT5-XXL). We employ beam search \cite{beamsearch} with a beam size of 4. All models are implemented with Huggingface \texttt{transformers} 
and evaluations are conducted using the \texttt{evaluate} library. 
All the experiments are conducted on a single Nvidia RTX8000 GPU with 49GB of GPU memory and 128GB of CPU memory for maximum 3 days on each experiment. All the results reported are based on a single run.

\subsection{Fine-Tuning Retrieval Augmented LMs for Personalization}

In the first sets of experiments, we establish baseline personalization results for a fine-tuned language model. We also investigate the impact of employing various retrieval techniques and the effect of retrieving different quantities of entries from a user profile. This analysis aims to provide insights into the efficacy of diverse retrieval methods and the potential benefits of adjusting the number of retrieved entries for personalization tasks.

\begin{table*}
    \centering
    \adjustbox{max width=\textwidth}{
        \begin{tabular}{llcccccccc}
        \toprule
         &  & \multicolumn{8}{c}{{FlanT5-base (fine-tuned)}} \\
         \cmidrule(lr){3-10}
        & & \multicolumn{2}{c}{{Non-Personalized}} & \multicolumn{4}{c}{{Untuned profile, $k = 1$}} & \multicolumn{2}{c}{{Tuned profile}} \\
        \cmidrule(lr){3-4} \cmidrule(lr){5-8} \cmidrule(lr){9-10}
         {{Dataset}} & {{Metric}} & No-Retrieval & Random & {Random} & {BM25} & {Contriever} & {Recency} & IPA & FiD($k = 16$) \\\midrule
        
        \multirow{2}{*}{\shortstack[l]{{\benchmark-1T: Personalized}\\{Citation Identification}}} & \multirow{2}{*}{Accuracy $\uparrow$} & \multirow{2}{*}{0.628} & \multirow{2}{*}{0.625} & \multirow{2}{*}{0.657} & \multirow{2}{*}{0.682} & \multirow{2}{*}{0.688} & \multirow{2}{*}{0.691} & \multirow{2}{*}{\textbf{0.714}} & \multirow{2}{*}{0.698} \\\\\midrule

        \multirow{2}{*}{\shortstack[l]{{\benchmark-2T: Personalized}\\{Movie Tagging}}} & Accuracy $\uparrow$ & 0.506 & 0.513 & 0.518 & 0.539 & 0.533 & 0.549 & {0.564} & \textbf{0.661} \\
        & F1 $\uparrow$ & 0.443 & 0.449 & 0.456 & 0.472 & 0.475 & 0.492 & {0.519} & \textbf{0.624} \\\midrule

        \multirow{2}{*}{\shortstack[l]{{\benchmark-3T: Personalized}\\{Product Rating}}} & MAE $\downarrow$ & 0.280 & 0.280 & 0.279 & 0.278 & 0.281 & 0.279 & {0.266} & \textbf{0.250} \\
        & RMSE $\downarrow$ & 0.615 & 0.616 & 0.612 & 0.614 & 0.606 & 0.608 & \textbf{0.598} & \textbf{0.598} \\\midrule

        \multirow{2}{*}{\shortstack[l]{{\benchmark-4T: Personalized}\\{News Headline Generation}}} & ROUGE-1 $\uparrow$ & 0.159 & 0.160 & 0.169 & 0.171 & 0.176 & 0.173 & \textbf{0.177} & 0.170 \\
        & ROUGE-L $\uparrow$ & 0.145 & 0.147 & 0.155 & 0.157 & 0.162 & 0.158 & \textbf{0.162} & 0.157 \\\midrule

        \multirow{2}{*}{\shortstack[l]{{\benchmark-5T: Personalized}\\{Scholarly Title Generation}}} & ROUGE-1 $\uparrow$ & 0.462 & 0.459 & 0.460 & 0.471 & 0.472 & 0.466 & \textbf{0.479} & 0.456 \\
        & ROUGE-L $\uparrow$ & 0.416 & 0.412 & 0.414 & 0.423 & 0.426 & 0.420 & \textbf{0.431} & 0.414 \\\midrule

        \multirow{2}{*}{\shortstack[l]{{\benchmark-6T: Personalized}\\{Email Subject Generation}}} & ROUGE-1 $\uparrow$ & 0.479 & 0.500 & 0.525 & 0.537 & 0.545 & 0.532 & \textbf{0.547} & 0.540 \\
        & ROUGE-L $\uparrow$ & 0.463 & 0.452 & 0.507 & 0.522 & 0.530 & 0.518 & \textbf{0.533} & 0.525 \\\midrule

        \multirow{2}{*}{\shortstack[l]{{\benchmark-7T: Personalized}\\{Tweet Paraphrasing}}} & ROUGE-1 $\uparrow$ & 0.462 & 0.474 & 0.505 & 0.508 & 0.505 & 0.503 & \textbf{0.516} & 0.502 \\
        & ROUGE-L $\uparrow$ & 0.416 & 0.457 & 0.456 & 0.457 & 0.455 & 0.453 & \textbf{0.465} & 0.450 \\
        \bottomrule
        \end{tabular}
    }
    \caption{The results for a fine-tuned LM on the test set of the time-based setting. The number of retrieved document for personalizing LM is denoted by $k$. Details for tuning the profile on validation sets is in Table \ref{tab:supervised-time-dev} in Appendix \ref{appendix:dev}.}
    \label{tab:supervised-time}
\end{table*}


\paragraph{Impact of Retrievers on Retrieval-Augmented Personalization Models.} 
Here, we study different implementations of $\mathcal{R}$ with a fine-tuned FlanT5-base model for generating personalized output: (1) a baseline random selector from the user profile, (2) BM25 \cite{Robertson1995OkapiBM25}, (3) Contriever\footnote{\url{https://huggingface.co/facebook/contriever}} \cite{contriever}, and (4) Recency, in which we select the latest item in the user profile based on time (only for time-based separation setting). BM25 is considered as a robust and strong term-matching retrieval model and Contriever is a pretrained dense retrieval model. 

The results of this experiment are shown in Table \ref{tab:supervised} for user-based separation and in Table \ref{tab:supervised-time} for time-based separation. The results suggest that personalization improves the performance for all tasks within the \benchmark benchmark. In most cases, even a random selection of documents from the user profile and the creation of personalized prompts leads to performance improvements compared to non-personalized prompts given to the LM. Note that non-personalized prompt can be achieved with no retrieval augmentation (No-Retrieval) or with augmentation with a random item from all user profiles. Results for more non-personalized baselines are presented in Appendix~\ref{appendix:other-baselines}.
\begin{figure*}[!t]
    \centering
    \includegraphics[width=\textwidth]{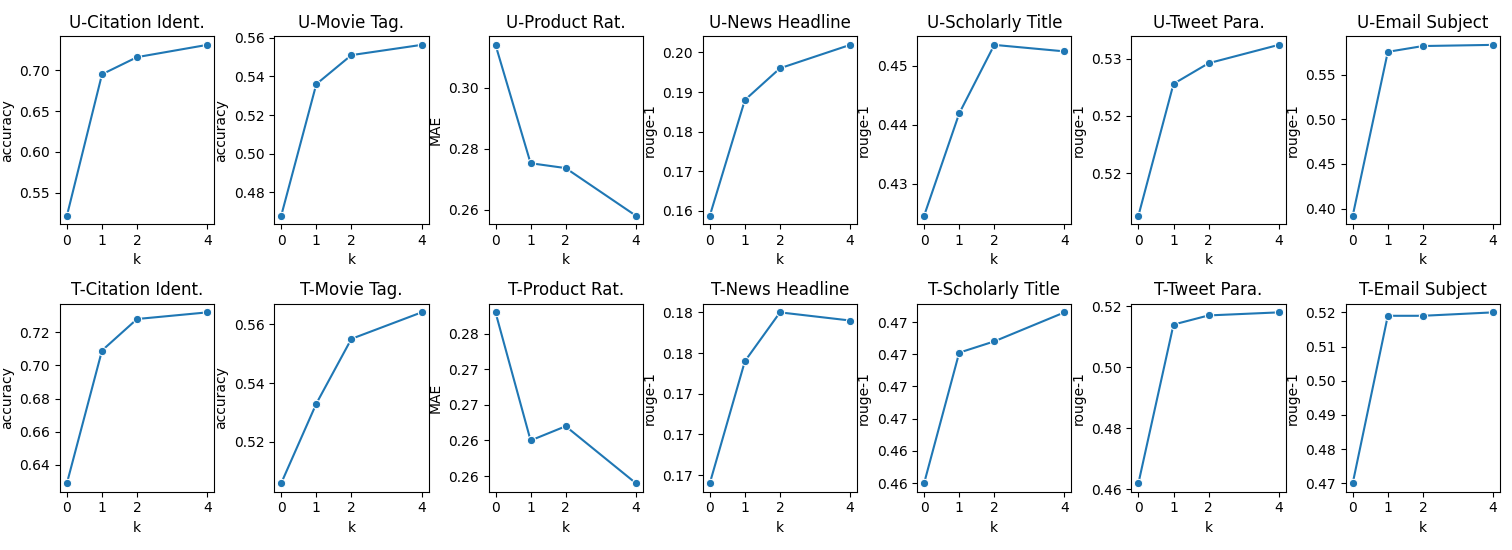}
    \caption{The performance on downstream tasks using the best retriever for each task from Tables \ref{tab:supervised} and \ref{tab:supervised-time} with different numbers of retrieved entries, $k$, from the user profile. The results indicate that increasing the number of retrieved documents for most datasets results in a better personalized performance.}
    \label{fig:ctx_effect}
\end{figure*}

When retrieving one document per user for personalizing the language model's output, Contriever demonstrates the best performance for most classification tasks (i.e., \benchmark-1U, \benchmark-2U, \benchmark-3U, \benchmark-1T, and \benchmark-2T). Recency only outperforms Contriever in the \benchmark-3T. Note that recency is considered as a simple yet strong personalization signal in search and recommendation \cite{recency-1, recency-2, recency-3}. For text generation, Contriever performs best for Personalization News Headline Generation (\benchmark-4U) and Personalized Tweet Paraphrasing (\benchmark-7U) in user-based separation setting. For Email Generation and Scholarly Title Generation tasks (\benchmark-5U and \benchmark-6U), BM25 demonstrates superior performance. Both BM25 and Contriever outperform a random profile selector in all \benchmark datasets. For the time-based separation setting, Contriever outperforms other methods in all generation tasks except News Headline Generation (\benchmark-4T), where recency performs better. 

Generally, the results indicate that incorporating any information from the user profile into the input is not sufficient, but rather selecting the most relevant and/or recent information is crucial. This underscores the importance of careful consideration in selecting and incorporating pertinent user profile elements in LLM prompts. 
There is no clear winner among the retrieval models we study and an ensemble of relevance and temporal signals for personalization should be studied in the future.

\paragraph{Impact of the Number of Retrieved Items, ${k}$, on LLM Personalization.} Each sample within this benchmark consists of a substantial number of user profile entries. As such, exploring the impact of incorporating multiple entries to augment the input of the language model can provide valuable insights into addressing the unresolved challenges posed by this benchmark. For the sake of space, we focus on our In-Prompt Augmentation (IPA) approach for personalization and depict the model's performance w.r.t different profile sizes in Figure \ref{fig:ctx_effect}. This experiment uses the best retriever from Tables \ref{tab:supervised} and \ref{tab:supervised-time} across various tasks, while varying the number of retrieved entries from user profiles. The results suggest that increasing the number of retrieved items leads to improved performance in downstream tasks. However, some tasks experience a decline in performance. Given the finite context size of language models, exploring approaches to generate a unified prompt from multiple user entries may represent promising future work. 

\paragraph{Impact of Tuning Retriever Hyperparameters.} 
Based on the performance on the validation set for each dataset, we tuned two parameters: (1) the retrieval model (BM25 vs. Contriever vs. Recency) for IPA and FiD, and (2) retrieved items count ($k$) for IPA. We consistently utilize 16 documents for FiD, as we do not observe much variance in the results. Both IPA and FiD approaches use FlanT5-base. For hyperparameter tuning, we used the following metrics on the development sets: Accuracy for \benchmark-1 and \benchmark-2, MAE for \benchmark-3, and ROUGE-1 for all text generation tasks. The results for this tuned model are presented in the last two columns of Table~\ref{tab:supervised} and Table~\ref{tab:supervised-time}. As expected, the tuned model outperforms the other models on all datasets. For text classification tasks, FiD surpasses the performance of IPA in all datasets, with the exception of LaMP-1T. Conversely, IPA exhibits superior performance across all text generation datasets.

\subsection{Zero-Shot Personalized Results for LLMs}
\begin{table*}[!t]
    \centering
    \adjustbox{max width=\textwidth}{
        \begin{tabular}{ll|cccc|cccc}
        \toprule
        & & \multicolumn{4}{|c}{User-based Separation} & \multicolumn{4}{|c}{Time-based Separation} \\
        \cmidrule(lr){3-6}\cmidrule(lr){7-10}
        & & \multicolumn{2}{|c}{{Non-Personalized}} & \multicolumn{2}{c}{{Personalized}} & \multicolumn{2}{|c}{{Non-Personalized}} & \multicolumn{2}{c}{{Personalized}} \\
        \cmidrule(lr){3-4}\cmidrule(lr){5-6}\cmidrule(lr){7-8}\cmidrule(lr){9-10}
        {{Dataset}}  & {{Metric}} & \multicolumn{1}{|c}{FlanT5-XXL} & {GPT-3.5} & {FlanT5-XXL} & {GPT-3.5} & \multicolumn{1}{|c}{FlanT5-XXL} & {GPT-3.5} & {FlanT5-XXL} & {GPT-3.5} \\\midrule
        
        \multirow{2}{*}{\shortstack[l]{{\benchmark-1: Personalized}\\{Citation Identification}}} & \multirow{2}{*}{Accuracy $\uparrow$} & \multirow{2}{*}{0.520} & \multirow{2}{*}{0.541} & \multirow{2}{*}{\textbf{0.699}} & \multirow{2}{*}{0.695} & \multirow{2}{*}{0.502} & \multirow{2}{*}{0.508} & \multirow{2}{*}{\textbf{0.636}} & \multirow{2}{*}{0.634} \\
        & & & & & & & & & \\\midrule

        \multirow{2}{*}{\shortstack[l]{{\benchmark-2: Personalized}\\{Movie Tagging}}} & Accuracy $\uparrow$ & 0.365 & 0.408 & 0.414 & \textbf{0.508} & 0.360 & 0.382 & 0.396 & \textbf{0.466} \\
        & F1 $\uparrow$ & 0.308 & 0.314 & 0.364 & \textbf{0.457} & 0.276 & 0.299 & 0.304 & \textbf{0.418} \\\midrule

        \multirow{2}{*}{\shortstack[l]{{\benchmark-3: Personalized}\\{Product Rating}}} & MAE $\downarrow$ & 0.344 & 0.706 & \textbf{0.267} & 0.620 & 0.333 & 0.677 & \textbf{0.299} & 0.603 \\
        & RMSE $\downarrow$ & 0.650 & 0.972 & \textbf{0.552} & 1.049 & 0.650 & 0.948 & \textbf{0.616} & 1.002 \\\midrule

        \multirow{2}{*}{\shortstack[l]{{\benchmark-4: Personalized}\\{News Headline Generation}}} & ROUGE-1 $\uparrow$ & 0.163 & 0.136 & \textbf{0.182} & 0.150 & 0.176 & 0.146 & \textbf{0.188} & 0.158 \\
        & ROUGE-L $\uparrow$ & 0.147 & 0.119 & \textbf{0.167} & 0.133 & 0.160 & 0.128 & \textbf{0.172} & 0.140 \\\midrule

        \multirow{2}{*}{\shortstack[l]{{\benchmark-5: Personalized}\\{Scholarly Title Generation}}} & ROUGE-1 $\uparrow$ & 0.442 & 0.387 & \textbf{0.450} & 0.390 & 0.471 & 0.424 & \textbf{0.483} & 0.425 \\
        & ROUGE-L $\uparrow$ & 0.400 & 0.329 & \textbf{0.411} & 0.329 & 0.422 & 0.355 & \textbf{0.433} & 0.351 \\\midrule

        \multirow{2}{*}{\shortstack[l]{{\benchmark-6: Personalized}\\{Email Subject Generation}}} & ROUGE-1 $\uparrow$ & 0.362 & - & \textbf{0.482} & - & 0.335 & - & \textbf{0.401} & - \\
        & ROUGE-L $\uparrow$ & 0.343 & - & \textbf{0.471} & - & 0.319 & - & \textbf{0.387} & - \\\midrule

        \multirow{2}{*}{\shortstack[l]{{\benchmark-7: Personalized}\\{Tweet Paraphrasing}}} & ROUGE-1 $\uparrow$ & \textbf{0.453} & 0.399 & 0.448 & 0.390 & \textbf{0.448} & 0.390 & 0.440 & 0.382 \\
        & ROUGE-L $\uparrow$ & \textbf{0.395} & 0.336 & 0.394 & 0.322 & \textbf{0.396} & 0.330 & 0.389 & 0.318 \\
        \bottomrule
        \end{tabular}
    }
    \caption{The zero-shot personalized results on the test set of user- and time-based separation settings. The tuned retriever was selected based on the validation performance as reported in Tables \ref{tab:llm-dev} and \ref{tab:llm-time-dev} in Appendix \ref{appendix:dev}. The results show that personalizing LLMs with the proposed approach improves all datasets in zero-shot setting except \benchmark-7.}
    \label{tab:llm}
\end{table*}
With the widespread adoption of employing LLMs with no fine-tuning in contemporary research, we conduct an evaluation of two such models on our benchmark.\footnote{As previously stated, FiD approach cannot be utilized with untrained models. Consequently, the experiments conducted in this section pertain solely to IPA method.} Particularly, we leverage GPT 3.5 (alias \texttt{gpt-3.5-turbo} or ChatGPT\footnote{\url{https://openai.com/blog/chatgpt}}) and FlanT5-XXL \cite{flant5}. FlanT5-XXL comprises 11B parameters, however, the size of GPT-3.5 is unknown (GPT3 consists of 175B parameters). For evaluation, we provide each model with the inputs corresponding to individual tasks and assess their performance based on the generated outputs. In classification tasks, if the produced output does not correspond to a valid class, we resort to calculating the similarity between each class label and the generated output utilizing BERTScore \cite{Zhang*2020BERTScore:}. Thus, we assign the most similar label to the generated output as the output for the given input. GPT-3.5 generated out-of-the-label predictions 8\%, 4\%, 6\%, 4\%, 2\%, and 4\% of the time for the LaMP-1U, LaMP-1T, LaMP-2U, LaMP-2T, LaMP-3U, and LaMP-3T tasks, respectively. On the other hand, FlanT5-XXL predictions are consistently among the questioned labels.

Table \ref{tab:llm} shows the result of LLMs on this benchmark in a zero-shot scenario. The results show that, except for the Personalized Tweet Paraphrasing task, using the user's profile with LLMs improves their performance on this benchmark in a zero-shot setting. The outcomes in Tables \ref{tab:supervised} and \ref{tab:supervised-time} show the results for FlanT5-base, a 250M parameter model, fine-tuned on each task. Table \ref{tab:llm} presents the zero-shot application of LLMs. These findings indicate that fine-tuning smaller models on downstream tasks leads to enhanced performance in comparison to zero-shot performance of LLMs.

Finally, it is crucial to highlight that the observed outcomes, which indicate superior performance of FlanT5-XXL over GPT-3.5, should not be construed as an inherent deficiency of the latter model. The efficacy of LLMs is extensively contingent upon the caliber and configuration of the input prompts. It is worth noting that prompt engineering, which plays a significant role in performance of LLMs, is not the central objective of this study. Consequently, any disparities in performance must be evaluated in light of this contextual information.

\section{Research Problems Enabled by \benchmark}

\benchmark can facilitate research in several areas, including but not limited to: 

\paragraph{Prompting Language Models for Personalization.} The integration of user profiles into language models can be approached using hard prompts, but their limited context size makes it difficult to include lengthy user profile entries. Exploring different prompts for personalization could be interesting. An alternative solution is generating personalization prompts based on the user profile, instead of relying on retrieved entries. Furthermore, the use of soft prompts \cite{soft-prompt} can be helpful for personalizing language models.

\paragraph{Evaluation of Personalized Text Generation.} The commonly used evaluation metrics for text generation, whether syntactical \cite{rouge, meteor, bleu} or semantical \cite{Zhang*2020BERTScore:}, do not incorporate the user into their evaluation process. Consequently, such metrics may not be entirely suitable for evaluating personalized text generation problems. Exploring new evaluation metrics that account for the user's preferences can benefit this area of research.

\paragraph{Learning to Retrieve from User Profiles.} Learning to rank has been widely explored in various retrieval scenarios. Optimizing ranking models that select personalized entries for the sake of personalized text classification and/or generation would be a potentially impactful research direction.

\section{Related Work}
Personalization has been well studied for information access problems, with the organization of the Netflix Challenge and its associated datasets representing an important driver of academic focus on personalization \cite{KonstanTerveen2021RecSys}. It also represents an important element of large-scale industry recommender systems \cite{davidson2010youtube, das2007googlenes, jiajing2022pinterest} and has also been extensively studied for search applications \cite{Bennett:2012,Dumais:2016, citeulike:2187446,PERSON:2018,Zeng:2023}, in contexts ranging from query auto-completion \cite{personalized-query} to collaborative personalized search \cite{10.1145/1462198.1462203}. We refer readers to \citet{rafieian2023ai} for an overview of this line of work. Here, we cover personalization in NLP, focusing on research datasets.

Personalization has been examined extensively for dialogue agents \cite{dialog2, dialog3, mazare2018training}. Compared to other NLP tasks. This focus likely stems from the importance of tailoring dialogue to users and conditioning generated utterances on specific personas. Given the lack of real conversational data, some work has constructed dialogue data for users by promoting crowd-workers to author dialogues based on specific personas \cite{dialog3}, and through extracting user attributes and utterances from Reddit \cite{mazare2018training, dialog2} and Weibo \cite{zhong2022less, Qian2021pchatbot}. To leverage more realistic conversational data, recent work of \citet{vincent2023personalised} annotate a dataset of movie dialogues with narrative character personas and posit the potential for using LLMs for dialogue generation conditioned on these personas. Other work has also leveraged publicly available reviews and recipes to explore personalization for review \cite{lituzhilin2019towards} and recipe generation \cite{majumder2019generating}. \citet{wuebker2018compact} explore parameter efficient models for personalized translation models. Finally, \citet{ao2021pens} presents a personalized headline generation dataset constructed from realistic user interaction data on Microsoft News. This is closely related to the \benchmark-4 task, which focuses on personalization for \emph{authors} rather than readers. \benchmark presents resources for the tasks which have seen lesser attention than those based on dialogue -- expanding the underexplored space of personalizing text classification/generation systems \cite{flek2020returning,dudy2021refocusing}.

While a body of work has focused on user-facing applications, others have explored personalization for more fundamental problems in language modeling. They have used openly available user data on Reddit \cite{personalizedLM2}, Facebook, Twitter \cite{soni2022human}, and other blogging websites \cite{PersonalizedLM1}. Besides pre-training LMs for personalization, \citet{soni2022human} explores applying a personalized LM for downstream tasks in stance classification and demographic inference. Similarly, other work has explored personalized sentiment prediction on publicly available Yelp and IMDB data \cite{mireshghallah2022useridentifier, zhong2021useradapter} -- this work bears a resemblance to the \benchmark-3 task and ties back to rating prediction explored in recommendation tasks. Finally, \citet{plepi2022unifying} examines the application of personalization methods to modeling annotators in a classification task reliant on modeling social norms -- making an important connection between personalization and an emerging body of work on accommodating human label variation in NLP \cite{rottger2022two, gordon2022jury, plank2022problem}.

\section{Conclusion}
This paper presented a novel benchmark named \benchmark for training and evaluating language models for personalized text classification and generation. \benchmark consists of seven datasets: three classification and four generation datasets. We proposed retrieval augmentation solutions for personalizing LLMs. Notably, we studied two augmentation approaches: in-prompt augmentation (IPA) and fusion-in-decoder (FiD). We performed extensive experiments using various LLMs and retrieval techniques for selecting user profile entries for producing personalized prompts. We demonstrated that our LLM personalization approaches can lead to 12.2\% average performance improvements across datasets on zero-shot setting, and 23.5\% with fine-tuning. Finally, we underscore the paramount importance of personalization in the current era dominated by large language models. We firmly believe that the future of natural language processing systems lies in a user-centric approach, tailoring solutions to individual needs for optimal effectiveness.

\section*{Acknowledgments}
This work was supported in part by the Center for Intelligent Information Retrieval, in part by NSF grant \#2143434, in part by the Office of Naval Research contract number N000142212688, in part by Google, Microsoft, and Lowe's. Any opinions, findings and conclusions or recommendations expressed in this material are those of the authors and do not necessarily reflect those of the sponsors.

\section*{Limitations}
\benchmark comes with certain limitations arising from task definitions, data leakage into LLM pre-training collections, effective evaluations for personalized generations and broader privacy concerns associated with personalization.

\paragraph{Task definitions} Although all tasks are designed to assess language models' proficiency in personalization, certain tasks could be better grounded in realistic scenarios and real-world applications. For instance, framing the Personalized Citation Identification task as a binary classification problem might not accurately represent real-world situations, where individuals generally need to interact with a more extensive array of articles. Additionally, while Personalized Product Rating is intrinsically linked to predicting user satisfaction, the approach may not be entirely realistic, as reviews in real-world contexts are often accompanied by a numerical rating, rendering direct score prediction less relevant. That being said, \benchmark creates an environment for evaluating the abilities of LMs in producing personalized outputs.

\paragraph{Leakage to LLM pretraining data} The data used for creating the \benchmark benchmark mostly consists of publicly available data on the Web, e.g., public tweets, scholarly articles, news articles, and product reviews. We should take this into consideration that some of this data may have been observed by the large language models during their pretraining process. Therefore, they may even perform poorer in unseen cases compared to what we observe from the results on most the \benchmark datasets. For this reason, we included the Avocado dataset for Personalized Email Subject Generation as this is not publicly available on the Web and we expect that language models do not use this dataset for pretraining given the restrictions on the data usage agreement. 

\paragraph{Evaluating personalized generation} The majority of text generation tasks addressed in this research employ short text generation as it offers greater convenience for evaluation purposes. Well-defined metrics, such as ROUGE and BLEU scores, are readily available to assess the quality of short text generation. Conversely, evaluating long text generation poses significant challenges due to its subjective nature, absence of a definitive reference, the necessity for coherence and consistency, contextual comprehension, varied output, semantic and factual accuracy, as well as the limitations of conventional metrics. Evaluators must account for multiple factors, encompassing structural integrity, clarity, effective employment of context, creativity, and subjectivity. Attaining consistent and objective evaluations proves arduous, as it heavily relies on human judgment, which can introduce biases.

\paragraph{Privacy and personalization} Finally, we urge the readers to be mindful of privacy implication of LLM personalization. Several studies have shown successful membership attacks against deep learning models (including LLMs) \cite{mattern2023membership,shokri2017}, thus using personal and private data in fine-tuning may put the privacy of some users at risk. Despite its importance, this paper does not study privacy issues and further analyses are required to ensure how fine-tuned personalized models can preserve the privacy of users. Note that we do not have privacy concerns for the proposed retrieval augmentation approach when a zero-shot language model is used, assuming that the zero-shot language model is hosted in-house or at a trusted third party.



\bibliography{anthology,custom}

\appendix

\section{Data Creation Details for Tasks in the \benchmark benchmark}
\label{appendix:lamp-data-creation}

This section explains the details behind creating inputs, outputs, and profile entries for each task in the \benchmark benchmark. Note that all datasets are in English language.

\subsection{User-based Separation Setting}

\paragraph{Personalized Citation Identification.} To generate data samples, we leverage the Citation Network Dataset (V14) \cite{Tang:08KDD}, which comprises information on scientific papers, authors, and citations. The dataset does not provide demographic details of the users in the data. We select all papers from this dataset that meet the following criteria: 1) they are written in English, 2) they contain at least one reference and one author, and 3) they include an abstract. Subsequently, we group papers based on their authors and only consider authors who have written at least 50 papers. For each author, we randomly select one of their papers and one of its cited references. For negative document selection, we randomly choose one of the first author's co-authors and one of the papers they have cited in one of their papers, which has not been cited by the first author. If no such author exists, we randomly select an author and repeat this process. Finally, we construct the input, output, and profile of the generated samples for this task, employing the template depicted in Figure \ref{fig:tasks_overview} in Appendix \ref{appendix:benchmark_samples}. After creating the samples for all users, we divide users into the train, validation, and test sets for this task.

\paragraph{Personalized Movie Tagging.} To construct our dataset for this task, we leverage the MovieLense dataset \cite{movie-lense} obtained from the MovieLense website\footnote{\url{https://movielens.org/}}. This dataset includes the tags that individual users have attributed to each film. However, it does not contain the descriptions or summaries of the films. To acquire these, we retrieve the overviews of each film in this dataset from the Movie Database website\footnote{\url{https://www.themoviedb.org/?language=en-US}}. The MovieLense dataset comprises over a thousand tags. However, for our dataset, we retain only the top 15 most frequently selected tags as labels. The dataset does not provide demographic details of the users in the data. Furthermore, for the sake of simplicity, we only include users who have assigned a single tag to a film. We only retain users with a minimum of five tagged movies. Then, we partition the users into training, validation, and test sets. For each movie tagged by a user, we use the movie description as input, the movie's tag as the output, and the remaining movies tagged by the same user and their tags as the user profile for that sample, following the template shown in Figure \ref{fig:tasks_overview} in Appendix \ref{appendix:benchmark_samples}. Finally, we randomly select 50\% of the generated samples for each user in training, validation, and test sets, and add them to the samples of the corresponding set.

\paragraph{Personalized Product Rating.} In this task, we create our dataset by leveraging the Amazon Reviews Dataset \cite{amazon-review}. We filtered out users (i.e., amazon customers who have written reviews) who have written less than $100$ and the 1\% users with the most reviews as outliers. Since the Amazon Reviews dataset is quite extensive, we randomly sampled a subset of users from the dataset, which we then split into training, validation, and testing sets. Note that the dataset does not provide demographic details of the users in the data.

To construct the input-output pairs for our task, for each user, we randomly select one of their reviews as the input to the task and use their other reviews as their profile. Specifically, we use the profile to capture the author's writing style, preferences, and tendencies. In this setup, the user's score for the input review serves as the ground truth output for our task. To gain a better understanding of the input, output, and profile, refer to Figure \ref{fig:tasks_overview} in Appendix \ref{appendix:benchmark_samples}.

\paragraph{Personalized News Headline Generation.} To construct our dataset for this task, we leverage the News Categorization dataset \cite{misra2022news, misra2021sculpting} from the HuffPost website. The dataset provides author information for each article and is used to group articles by their respective authors. The dataset does not provide demographic details of the users in the data. We filtered out the authors with less than four articles. In cases where an article has multiple authors, we assign it only to the first author.

We then randomly split the authors into training, validation, and test sets. For each author in each set, we create input-output pairs by selecting each article as the input, the headline of the article as the output, and the remaining articles written by the same author as their profile. This setup aims to capture the author's writing style, preferences, and tendencies, which can be leveraged to generate headlines that align with their interests. An example of this setup is presented in Figure \ref{fig:tasks_overview} in Appendix \ref{appendix:benchmark_samples}. Finally, to ensure a diverse and representative dataset, we randomly select 50\% of the created samples for each author and add them to the user's corresponding set. 

\paragraph{Personalized Scholarly Title Generation.} Similar to Section \benchmark-1, we leverage the Citation Network Dataset (V14) \cite{Tang:08KDD} that includes information about scientific papers, authors, and citations to construct our dataset. The dataset does not provide demographic details of the users in the data. We only kept the papers that meet the following criteria: 1) written in English, 2) have at least one reference and one author, and 3) have an abstract. Then, we group papers by their authors and only consider authors who have published at least 50 papers. For each author, we randomly choose one of their papers and use its abstract as input, its title as output, and the remaining papers as the author's profile. Figure \ref{fig:tasks_overview} in Appendix \ref{appendix:benchmark_samples} illustrates the input format for this task. After creating the samples for all users, we divide users into the train, validation, and test sets for this task.

\paragraph{Personalized Email Subject Generation.} In this study, we adopt the Avocado Research Email Collection \cite{avocado} as the primary dataset for our task. The dataset does not provide demographic details of the users in the data. To curate the dataset, we first perform a filtering step where we exclude emails with subject lengths of fewer than five words and content lengths of fewer than 30 words. Next, we group the emails based on their sender's email address, retaining only those from users with email frequencies ranging between 10 to 200 emails. We further divide the users into distinct training, validation, and test sets to ensure that our model generalizes well to unseen data. To generate training examples for each user, we create input-output pairs by considering each email as the input and the corresponding email subject as the output. We supplement these pairs with other emails written by the same user as their profile, as shown in Figure \ref{fig:tasks_overview} in Appendix \ref{appendix:benchmark_samples}. We ensure that our dataset is diverse and representative by randomly selecting 50\% of the curated samples for each user and adding them to their respective sets. 

\paragraph{Personalized Tweet Paraphrasing.} In this task, we utilize the Sentiment140 dataset \cite{Sentiment140} as our tweet collection set. The dataset does not provide demographic details of the users in the data. To ensure that the collected tweets are of adequate length, we only retain tweets containing at least 10 words. We then group the tweets based on the user ID and filter out users with fewer than 10 tweets. Subsequently, we randomly select one tweet from each user profile and use it as input to ChatGPT (i.e., \texttt{gpt3.5-turbo}) to generate a paraphrased version. The generated paraphrase is then utilized as the input to our NLP task, with the original tweet serving as the corresponding output. The remaining tweets of the user constitute the user's profile, excluding the one selected as input. Figure \ref{fig:tasks_overview} in Appendix \ref{appendix:benchmark_samples} provides an overview of the input-output-profile template for our proposed task. After creating the samples for all users, we divide users into the train, validation, and test sets for this task.

\subsection{Time-based Separation Setting}

\paragraph{Personalized Citation Identification.} To generate data samples, we leverage the Citation Network Dataset (V14) \cite{Tang:08KDD}, which comprises information on scientific papers, authors, and citations. The dataset does not provide demographic details of the users in the data. We select all papers from this dataset that meet the following criteria: 1) they are written in English, 2) they contain at least one reference and one author, and 3) they include an abstract. Subsequently, we group papers based on their authors and only consider authors who have written at least 50 papers. We divide each author's papers based on the publication year into three groups chronologically: 1) profile papers, 2) train papers, 3) validation papers, and 4) test papers, where the order of groups shows the flow of time. Each train, validation, and test paper set in this task consists of only one paper. Then, for each paper in the train, validation, and test sets for the user, we select each paper and one of its cited references. For negative document selection, we randomly choose one of the first author's co-authors and one of the papers they have cited in one of their papers, which has not been cited by the first author. If no such author exists, we randomly select an author and repeat this process. Finally, we construct the input, output, and profile of the generated samples for this task, employing the template depicted in Figure \ref{fig:tasks_overview} in Appendix \ref{appendix:benchmark_samples}. It should be noted that after creating all the samples in the train, validation, and test sets for all the users and aggregating them, we randomly select a subset of validation and test sets to create the final sets for the task.

\paragraph{Personalized Movie Tagging.} 

To construct our dataset for this task, we leverage the MovieLense dataset \cite{movie-lense} obtained from the MovieLense website\footnote{\url{https://movielens.org/}}. This dataset includes the tags that individual users have attributed to each film. The dataset does not provide demographic details of the users in the data. However, it does not contain the descriptions or summaries of the films. To acquire these, we retrieve the overviews of each film in this dataset from the Movie Database website\footnote{\url{https://www.themoviedb.org/?language=en-US}}. The MovieLense dataset comprises over a thousand tags. However, for our dataset, we retain only the top 15 most frequently selected tags as labels. Furthermore, for the sake of simplicity, we only include users who have assigned a single tag to a film. We only retain users with a minimum of five tagged movies. We divide each user's tagged movies based on the date they tagged into three groups, chronologically: 1) profile movies, 2) train movies, 3) validation movies, and 4) test movies, where the order of groups shows the flow of time. Each train, validation, and test movies set in this task consists of 20\%, 10\%, and 10\% of movies, respectively. Then, for each movie in the train, validation, and test sets for the user, we use the movie's description as the input, the movie's tag as the output, and the profile movies and their tags as the profile for that sample, following the template shown in Figure \ref{fig:tasks_overview} in Appendix \ref{appendix:benchmark_samples}.  It should be noted that after creating all the samples in the train, validation, and test sets for all the users and aggregating them, we randomly select a subset of validation and test sets to create the final sets for the task.

\paragraph{Personalized Product Rating.} In this task, we create our dataset by leveraging the Amazon Reviews Dataset \cite{amazon-review}. The dataset does not provide demographic details of the users in the data. We filtered out users (i.e., amazon customers who have written reviews) who have written less than $100$ and the 1\% users with the most reviews as outliers. Since the Amazon Reviews dataset is quite extensive, we randomly sampled a subset of users from the dataset. We divide each user's reviews based on the review date into three groups chronologically: 1) profile reviews, 2) train reviews, 3) validation reviews, and 4) test reviews, where the order of groups shows the flow of time. Each train, validation, and test reviews set in this task consists of only one review.

To construct the input-output pairs for our task, for each user, we select their reviews in each of train, validation, and test sets as the input to the task and use the profile reviews as their profile. Specifically, we use the profile to capture the author's writing style, preferences, and tendencies. Additionally, the user's score for the input review serves as the ground truth output for our task. To gain a better understanding of the input, output, and profile, refer to Figure \ref{fig:tasks_overview} in Appendix \ref{appendix:benchmark_samples}. It should be noted that after creating all the samples in the train, validation, and test sets for all the users and aggregating them, we randomly select a subset of validation and test sets to create the final sets for the task.

\paragraph{Personalized News Headline Generation.} To construct our dataset for this task, we leverage the News Categorization dataset \cite{misra2022news, misra2021sculpting} from the HuffPost website. The dataset provides author information for each article and is used to group articles by their respective authors. The dataset does not provide demographic details of the users in the data. We filtered out the authors with less than ten articles. In cases where an article has multiple authors, we assign it only to the first author. We divide each author's articles based on the publishing date into three groups chronologically: 1) profile articles, 2) train articles, 3) validation articles, and 4) test articles, where the order of groups shows the flow of time. Each train, validation, and test articles set in this task consists of 20\%, 10\%, and 10\% articles, respectively. Then, for each article in the train, validation, and test sets for the user, we create input-output pairs by selecting each article as the input, the headline of the article as the output, and the profile articles written by the same author as their profile. This setup aims to capture the author's writing style, preferences, and tendencies, which can be leveraged to generate headlines that align with their interests. An example of this setup is presented in Figure \ref{fig:tasks_overview} in Appendix \ref{appendix:benchmark_samples}. It should be noted that after creating all the samples in the train, validation, and test sets for all the users and aggregating them, we randomly select a subset of validation and test sets to create the final sets for the task.

\paragraph{Personalized Scholarly Title Generation.} We leverage the Citation Network Dataset (V14) \cite{Tang:08KDD} that includes information about scientific papers, authors, and citations to construct our dataset. The dataset does not provide demographic details of the users in the data. We only kept the papers that meet the following criteria: 1) written in English, 2) have at least one reference and one author, and 3) have an abstract. Then, we group papers by their authors and only consider authors who have published at least 50 papers. We divide each author's papers based on the publication year into three groups chronologically: 1) profile papers, 2) train papers, 3) validation papers, and 4) test papers, where the order of groups shows the flow of time. Each train, validation, and test paper set in this task consists of only one paper. Then, for each paper in the train, validation, and test sets for the user, we use its abstract as input, its title as output, and the profile papers as the author's profile. Figure \ref{fig:tasks_overview} in Appendix \ref{appendix:benchmark_samples} illustrates the input format for this task. After creating the samples for all users, we divide users into the train, validation, and test sets for this task. It should be noted that after creating all the samples in the train, validation, and test sets for all the users and aggregating them, we randomly select a subset of validation and test sets to create the final sets for the task.

\paragraph{Personalized Email Subject Generation.} In this study, we adopt the Avocado Research Email Collection \cite{avocado} as the primary dataset for our task. The dataset does not provide demographic details of the users in the data. To curate the dataset, we first perform a filtering step where we exclude emails with subject lengths of fewer than five words and content lengths of fewer than 30 words. Next, we group the emails based on their sender's email address, retaining only those from users with email frequencies ranging between 10 to 200 emails. We divide each user's emails based on the publishing date into three groups chronologically: 1) profile emails, 2) train emails, 3) validation emails, and 4) test emails, where the order of groups shows the flow of time. Each train, validation, and test emails set in this task consists of 20\%, 10\%, and 10\% articles, respectively. Then, for each article in the train, validation, and test sets for the user, we create input-output pairs by considering each email as the input and the corresponding email subject as the output. We supplement these pairs with profile emails written by the same user as their profile, as shown in Figure \ref{fig:tasks_overview} in Appendix \ref{appendix:benchmark_samples}. It should be noted that after creating all the samples in the train, validation, and test sets for all the users and aggregating them, we randomly select a subset of validation and test sets to create the final sets for the task.

\paragraph{Personalized Tweet Paraphrasing.} In this task, we utilize the Sentiment140 dataset \cite{Sentiment140} as our tweet collection set. The dataset does not provide demographic details of the users in the data. To ensure that the collected tweets are of adequate length, we only retain tweets containing at least 10 words. We then group the tweets based on the user ID and filter out users with fewer than 10 tweets. We divide each user's tweets based on the publication year into three groups chronologically: 1) profile tweets, 2) train tweets, 3) validation tweets, and 4) test tweets, where the order of groups shows the flow of time. Each train, validation, and test tweet set in this task consists of only one paper. Then, for each tweet in the train, validation, and test sets for the user, we use it as input to ChatGPT (i.e., \texttt{gpt3.5-turbo}) to generate a paraphrased version. The generated paraphrase is then utilized as the input to our NLP task, with the original tweet serving as the corresponding output. The profile tweets of the user constitute the user's profile. Figure \ref{fig:tasks_overview} in Appendix \ref{appendix:benchmark_samples} provides an overview of the input-output-profile template for our proposed task. It should be noted that after creating all the samples in the train, validation, and test sets for all the users and aggregating them, we randomly select a subset of validation and test sets to create the final sets for the task.

\section{Samples of the Tasks Introduced in the \benchmark Benchmark}
\label{appendix:benchmark_samples}

\begin{figure*}[!t]
    \centering
    \includegraphics[width=0.83\textwidth]{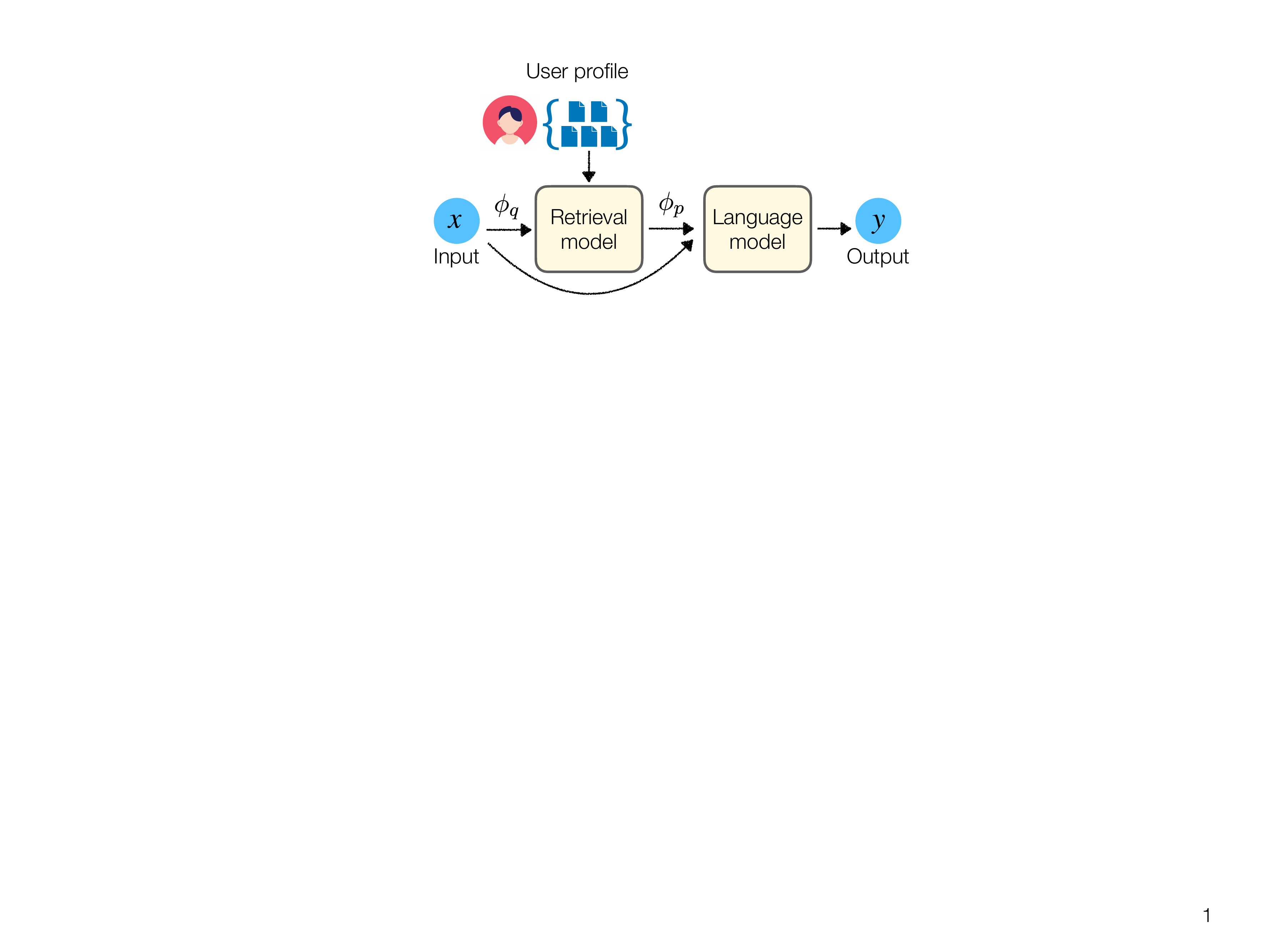}
    \caption{An overview of the templates used for creating data samples for each task in \benchmark. \texttt{Teletype} text is replaced with realistic data for each task.}
    \label{fig:tasks_overview}
\end{figure*}

As mentioned earlier, \benchmark proposes seven tasks to evaluate language model personalization. In order to create the data points, we use just a carefully designed template for each task. Figure \ref{fig:tasks_overview} depicts a sample and template for each task in \benchmark. Generally, each sample in each task has an input and output accompanied by a profile consisting of several entries about the user, helping the model to produce personalized results for the user. While the profile entries in the same task have a similar structure, the structure varies between tasks. For example, Figure \ref{fig:tasks_overview} shows that the profile for Personalized Product Rating comprised of documents with text and score sections, while the profile entries in Personalized Scholarly Title Generation have abstract and title attributes.

\section{Prompts Used for Adding User Profile to the Language Model's Input}
\label{appendix:prompts-creation}

\begin{table*}[!t]
    \centering
    \begin{adjustbox}{max width=\textwidth}
    \begin{tabular}{p{4cm}p{7cm}p{9cm}}
        \textbf{Task} & \textbf{Per Profile Entry Prompt (\texttt{PPEP})} & \textbf{Aggregated Input Prompt(AIP)} \\
        \toprule
            1: Citation Identification & ``$P_i$\texttt{\texttt{[title]}}'' & \textcolor{blue}{\texttt{add\_to\_paper\_title}}(\textcolor{blue}{\texttt{concat}}([\textcolor{blue}{\texttt{PPEP}}($P_1$), ..., \textcolor{blue}{\texttt{PPEP}}($P_n$)], \textcolor{gray}{", and "}), \textcolor{red}{\texttt{[INPUT]}}) \\
            2: Movie Tagging & the tag for the movie: ``$P_i$\texttt{[description]}'' is ``$P_i$\texttt{[tag]}'' & \textcolor{blue}{\texttt{concat}}([\textcolor{blue}{\texttt{PPEP}}($P_1$), ..., \textcolor{blue}{\texttt{PPEP}}($P_n$)], \textcolor{gray}{``, and ''}). \textcolor{red}{\texttt{[INPUT]}} \\
            3: Product Rating & $P_i$[score] is the score for ``$P_i$\texttt{[text]}'' & \textcolor{blue}{\texttt{concat}}([\textcolor{blue}{\texttt{PPEP}}($P_1$), ..., \textcolor{blue}{\texttt{PPEP}}($P_n$)], \textcolor{gray}{``, and ''}). \textcolor{red}{\texttt{[INPUT]}} \\
            4: News Headline Generation & ``$P_i$\texttt{[title]}'' is the title for ``$P_i$\texttt{[text]}'' & \textcolor{blue}{\texttt{concat}}([\textcolor{blue}{\texttt{PPEP}}($P_1$), ..., \textcolor{blue}{\texttt{PPEP}}($P_n$)], \textcolor{gray}{``, and ''}). \textcolor{red}{\texttt{[INPUT]}} \\
            5: Scholarly Title Generation & ``$P_i$\texttt{[title]}'' is the title for ``$P_i$[abstract]'' & \textcolor{blue}{\texttt{concat}}([\textcolor{blue}{\texttt{PPEP}}($P_1$), ..., \textcolor{blue}{\texttt{PPEP}}($P_n$)], \textcolor{gray}{``, and ""})\textcolor{gray}{. Following the given patterns} \textcolor{red}{\texttt{[INPUT]}} \\
            6: Email Subject Generation & ``$P_i$\texttt{[title]}'' is the title for ``$P_i$\texttt{[text]}'' & \textcolor{blue}{\texttt{concat}}([\textcolor{blue}{\texttt{PPEP}}($P_1$), ..., \textcolor{blue}{\texttt{PPEP}}($P_n$)], \textcolor{gray}{``, and ''}). \textcolor{red}{\texttt{[INPUT]}} \\
            7: Tweet Paraphrasing & ``$P_i$\texttt{[text]}'' & \textcolor{blue}{\texttt{concat}}([\textcolor{blue}{\texttt{PPEP}}($P_1$), ..., \textcolor{blue}{\texttt{PPEP}}($P_n$)], \textcolor{gray}{``, and ''}) \textcolor{gray}{are written by a person. Following the given patterns} \textcolor{red}{\texttt{[INPUT]}}\\
            \bottomrule
    \end{tabular}
    \end{adjustbox}
    \caption{Prompts template used to augment the input of the LM with the user profile. \textcolor{blue}{\texttt{concat}} is a function that \texttt{concat}enates the strings in its first argument by placing the string in the second argument between them. \textcolor{blue}{\texttt{add\_to\_paper\_title}} is a function designed to add the string in its first argument to the paper's title in the Personalized Citation Identification task. \textcolor{blue}{\texttt{PPEP}} is a function that create the prompt for each entry in the retrieved profile entries. \textcolor{red}{\texttt{[INPUT]}} is the task's input.}
    \label{tab:task-prompts}
\end{table*}

In order to use multiple entries from the user profile to personalize the language model's input, we construct task-specific prompts using the templates and instructions in Table \ref{tab:task-prompts}.

The prompt creation consists of two stages: 1) Per Profile Entry Prompt (PPEP) creation and 2) Aggregated Input Prompt (AIP) creation. In the first stage, following the instructions in Table \ref{tab:task-prompts}, we create a prompt for each profile entry. In the second stage, following the instructions in Table \ref{tab:task-prompts}, we combine the PPEP prompts into a single prompt to be fed to the language model. It should be noted that due to the limited context size of language models, we need to trim the PPEP prompts. More accurately, considering $k$ prompts need to be merged and that the maximum capacity for the task input is $\bar{L}$ and the maximum context size of the language model is $L$, we let each PPEP occupy $\frac{L-\bar{L}}{k}$ tokens in the language model's input. For PPEPs that are longer than the calculated number, we trim the non-template parts that have less importance in the final performance of the model -- the parts that do not provide category, score, or title. We select $\bar{L} = 256$ in this paper.



\section{Performance of the Models on the Validation Set}
\label{appendix:dev}

This section reports the results of experiments on the validation set. Table \ref{tab:supervised-dev} reports the results of fine-tuning the language model on the user-based separation setting on the validation set. Table \ref{tab:supervised-time-dev} shows the results of fine-tuning the language model on the time-based separation setting on the validation set. Table \ref{tab:llm-dev} shows the results of zero-shot evaluation of large language models on the user-based separation setting on the validation set. Table \ref{tab:llm-time-dev} depicts the results of zero-shot evaluation of large language models on the time-based separation setting on the validation set.

\begin{table*}[!bt]
    \centering
    \scalebox{0.69}{
        \begin{tabular}{llcccccccc}
        \toprule
         & & \multicolumn{6}{c}{{FlanT5-base (fine-tuned)}} \\
         \cmidrule(lr){3-10}
        & & \multicolumn{2}{c}{{Non-Personalized}} & \multicolumn{3}{c}{{Untuned profile, $k = 1$}} & {{Tuned retriever, $k$}} & \multicolumn{2}{c}{{Tuned profile}} \\
        \cmidrule(lr){3-4} \cmidrule(lr){5-7} \cmidrule(lr){8-8} \cmidrule(lr){9-10}
         {{Dataset}} & {{Metric}} & No-Retrieval & Random  & {Random} & {BM25} & {Contriever} & & IPA & FiD($k = 16$) \\\midrule
    
        \multirow{2}{*}{\shortstack[l]{{\benchmark-1U: Personalized}\\{Citation Identification}}} & \multirow{2}{*}{Accuracy $\uparrow$} & \multirow{2}{*}{0.522} & \multirow{2}{*}{0.526} & \multirow{2}{*}{0.597} & \multirow{2}{*}{0.623} & \multirow{2}{*}{0.695} & \multirow{2}{*}{Contriever, 4} & \multirow{2}{*}{{0.731}} & \multirow{2}{*}{\textbf{0.743}}  \\ \\\midrule

        \multirow{2}{*}{\shortstack[l]{{\benchmark-2U: Personalized}\\{Movie Tagging}}} & Accuracy $\uparrow$ & 0.449 & 0.447 & 0.513 & 0.498 & 0.524 & \multirow{2}{*}{Contriever, 4} & {0.560} & \textbf{0.640} \\
        & F1 $\uparrow$ & 0.403 & 0.405 & 0.462 & 0.442 & 0.472 & & {0.512} & \textbf{0.613} \\\midrule

        \multirow{2}{*}{\shortstack[l]{{\benchmark-3U: Personalized}\\{Product Rating}}} & MAE $\downarrow$ & 0.314 & 0.324 & 0.312 & 0.282 & 0.275 & \multirow{2}{*}{Contriever, 4} & \textbf{0.258} & 0.259 \\
        & RMSE $\downarrow$ & 0.624 & 0.650  & 0.633 & 0.609 & 0.589 & & \textbf{0.572} & 0.577 \\\midrule

        \multirow{2}{*}{\shortstack[l]{{\benchmark-4U: Personalized}\\{News Headline Generation}}} & ROUGE-1 $\uparrow$ & 0.158 & 0.163  & 0.167 & 0.176 & 0.188 & \multirow{2}{*}{Contriever, 4} & \textbf{0.201} & 0.194 \\
        & ROUGE-L $\uparrow$ & 0.144 & 0.151 & 0.152 & 0.161 & 0.172 & & \textbf{0.185} & 0.180 \\\midrule

        \multirow{2}{*}{\shortstack[l]{{\benchmark-5U: Personalized}\\{Scholarly Title Generation}}} & ROUGE-1 $\uparrow$ & 0.424 & 0.387 & 0.389 & 0.441 & 0.405 & \multirow{2}{*}{BM25, 2} & \textbf{0.453} & 0.445 \\
        & ROUGE-L $\uparrow$ & 0.382 & 0.350 & 0.352 & 0.401 & 0.367 & & \textbf{0.414} & 0.405 \\\midrule

        \multirow{2}{*}{\shortstack[l]{{\benchmark-6U: Personalized}\\{Email Subject Generation}}} & ROUGE-1 $\uparrow$ & 0.392 & 0.466 & 0.469 & 0.575 & 0.567 & \multirow{2}{*}{BM25, 4} & \textbf{0.583} & 0.559 \\
        & ROUGE-L $\uparrow$ & 0.374 & 0.452 & 0.454 & 0.563 & 0.553 & & \textbf{0.570} & 0.547 \\\midrule

        \multirow{2}{*}{\shortstack[l]{{\benchmark-7U: Personalized}\\{Tweet Paraphrasing}}} & ROUGE-1 $\uparrow$ & 0.511 & 0.512 & 0.512 & 0.520 & 0.522 & \multirow{2}{*}{Contriever, 4} & \textbf{0.526} & 0.511 \\
        & ROUGE-L $\uparrow$ & 0.456 & 0.456 & 0.457 & 0.465 & 0.467 & & \textbf{0.471} & 0.457 \\
        \bottomrule
        \end{tabular}
    }
    \caption{The personalized text classification and generation results for a fine-tuned language model (i.e., FlanT5-base) on the validation set of user-based separation setting. $k$ denotes the number of documents retrieved for personalizing language model outputs.}
    \label{tab:supervised-dev}
\end{table*}

\begin{table*}[!bt]
    \centering
    \scalebox{1}{
        \begin{tabular}{llcccc}
        \toprule
         & & \multicolumn{2}{c}{{Non-Personalized}} & \multicolumn{2}{c}{{Personalized}} \\
        \cmidrule(lr){3-4}\cmidrule(lr){5-6}
        {{Dataset}} & {{Metric}} & {FlanT5-XXL} & {GPT-3.5} & {FlanT5-XXL} & {GPT-3.5} \\\midrule
        
        \multirow{2}{*}{\shortstack[l]{{\benchmark-1U: Personalized}\\{Citation Identification}}} & \multirow{2}{*}{Accuracy $\uparrow$} & \multirow{2}{*}{0.522} & \multirow{2}{*}{0.510} & \multirow{2}{*}{0.675} & \multirow{2}{*}{\textbf{0.701}} \\\\\midrule

        \multirow{2}{*}{\shortstack[l]{{\benchmark-2U: Personalized}\\{Movie Tagging}}} & Accuracy $\uparrow$ & 0.348 & 0.372 & 0.369 & \textbf{0.466} \\
        & F1 $\uparrow$ & 0.268 & 0.290 & 0.294 & \textbf{0.424} \\\midrule
        
        \multirow{2}{*}{\shortstack[l]{{\benchmark-3U: Personalized}\\{Product Rating}}} & MAE $\downarrow$ & 0.357 & 0.699 & \textbf{0.282} & 0.658 \\
        & RMSE $\downarrow$ & 0.666 & 0.977 & \textbf{0.5841} & 1.102 \\\midrule

        \multirow{2}{*}{\shortstack[l]{{\benchmark-4U: Personalized}\\{News Headline Generation}}} & ROUGE-1 $\uparrow$ & 0.164 & 0.133 & \textbf{0.192} & 0.160 \\
        & ROUGE-L $\uparrow$ & 0.149 & 0.118 & \textbf{0.178} & 0.142 \\\midrule

        \multirow{2}{*}{\shortstack[l]{{\benchmark-5U: Personalized}\\{Scholarly Title Generation}}} & ROUGE-1 $\uparrow$ & 0.455 & 0.395 & \textbf{0.467} & 0.398 \\
        & ROUGE-L $\uparrow$ & 0.410 & 0.334 & \textbf{0.424} & 0.336 \\\midrule

        \multirow{2}{*}{\shortstack[l]{{\benchmark-6U: Personalized}\\{Email Subject Generation}}} & ROUGE-1 $\uparrow$ & 0.332 & - & \textbf{0.466} & - \\
        & ROUGE-L $\uparrow$ & 0.320 & - & \textbf{0.453} & - \\\midrule

        \multirow{2}{*}{\shortstack[l]{{\benchmark-7U: Personalized}\\{Tweet Paraphrasing}}} & ROUGE-1 $\uparrow$ & \textbf{0.459} & 0.396 & 0.448 & 0.391 \\
        & ROUGE-L $\uparrow$ & \textbf{0.404} & 0.337 & 0.396 & 0.324 \\
        \bottomrule
        \end{tabular}
    }
    \caption{The zero-shot personalized text classification and generation results on the validation set of user-based separation setting. For personalized models, the tuned retriever based on the validation performance was selected.}
    \label{tab:llm-dev}
\end{table*}

\begin{table*}
    \centering
    \scalebox{0.63}{
        \begin{tabular}{llccccccccc}
        \toprule
         & & \multicolumn{8}{c}{{FlanT5-base (fine-tuned)}} \\
         \cmidrule(lr){3-11}
        & & \multicolumn{2}{c}{{Non-Personalized}} & \multicolumn{4}{c}{{Untuned profile, $k = 1$}} & {{Tuned retriever, k}} & \multicolumn{2}{c}{{Tuned profile}} \\
        \cmidrule(lr){3-4} \cmidrule(lr){5-8} \cmidrule(lr){9-9} \cmidrule(lr){10-11}
         {{Dataset}} & {{Metric}} & No-Retrieval & Random & {Random} & {BM25} & {Contriever} & {Recency} & & IPA & FiD($k = 16$) \\\midrule
        
        \multirow{2}{*}{\shortstack[l]{{\benchmark-1T: Personalized}\\{Citation Identification}}} & \multirow{2}{*}{Accuracy $\uparrow$} & \multirow{2}{*}{0.629} & \multirow{2}{*}{0.630} & \multirow{2}{*}{0.662} & \multirow{2}{*}{0.695} & \multirow{2}{*}{0.709} & \multirow{2}{*}{0.681} & \multirow{2}{*}{Contriever, 4} & \multirow{2}{*}{\textbf{0.732}} & \multirow{2}{*}{0.694} \\\\\midrule

        \multirow{2}{*}{\shortstack[l]{{\benchmark-2T: Personalized}\\{Movie Tagging}}} & Accuracy $\uparrow$ & 0.512 & 0.506 & 0.531 & 0.538 & 0.551 & 0.546 & \multirow{2}{*}{Contriever, 4} & {0.570} & \textbf{0.658} \\
        & F1 $\uparrow$ & 0.460 & 0.453 & 0.480 & 0.485 & 0.505 & 0.493 & & {0.522} & \textbf{0.615} \\\midrule

        \multirow{2}{*}{\shortstack[l]{{\benchmark-3T: Personalized}\\{Product Rating}}} & MAE $\downarrow$ & 0.278 & 0.276 & 0.273 & 0.269 & 0.272 & 0.260 & \multirow{2}{*}{Recency, 4} & {0.259} & \textbf{0.252} \\
        & RMSE $\downarrow$ & 0.595 & 0.598 & 0.590 & 0.583 & 0.589 & 0.576 & & \textbf{0.568} & 0.586 \\\midrule

        \multirow{2}{*}{\shortstack[l]{{\benchmark-4T: Personalized}\\{News Headline Generation}}} & ROUGE-1 $\uparrow$ & 0.164 & 0.166 & 0.176 & 0.176 & 0.177 & 0.179 & \multirow{2}{*}{Recency, 2} & \textbf{0.185} & 0.178 \\
        & ROUGE-L $\uparrow$ & 0.149 & 0.151 & 0.160 & 0.161 & 0.163 & 0.165 & & \textbf{0.169} & 0.164 \\\midrule

        \multirow{2}{*}{\shortstack[l]{{\benchmark-5T: Personalized}\\{Scholarly Title Generation}}} & ROUGE-1 $\uparrow$ & 0.462 & 0.458 & 0.459 & 0.473 & 0.470 & 0.462 & \multirow{2}{*}{Contriever, 4} & \textbf{0.472} & 0.454 \\
        & ROUGE-L $\uparrow$ & 0.414 & 0.410 & 0.412 & 0.425 & 0.423 & 0.416 & & \textbf{0.423} & 0.411 \\\midrule

        \multirow{2}{*}{\shortstack[l]{{\benchmark-6T: Personalized}\\{Email Subject Generation}}} & ROUGE-1 $\uparrow$ & 0.470 & 0.503 & 0.504 & 0.509 & 0.519 & 0.510 & \multirow{2}{*}{Contriever, 4} & \textbf{0.520} & 0.513 \\
        & ROUGE-L $\uparrow$ & 0.455 & 0.450 & 0.489 & 0.496 & 0.507 & 0.497 & & \textbf{0.509} & 0.500 \\\midrule

        \multirow{2}{*}{\shortstack[l]{{\benchmark-7T: Personalized}\\{Tweet Paraphrasing}}} & ROUGE-1 $\uparrow$ & 0.462 & 0.462 & 0.507 & 0.509 & 0.514 & 0.510 & \multirow{2}{*}{Contriever, 4} & \textbf{0.518} & 0.505 \\
        & ROUGE-L $\uparrow$ & 0.414 & 0.448 & 0.457 & 0.460 & 0.464 & 0.459 & & \textbf{0.467} & 0.455 \\
        \bottomrule
        \end{tabular}
    }
    \caption{The personalized text classification and generation results for a fine-tuned language model (i.e., FlanT5-base) on the validation set of time-based separation setting. $k$ denotes the number of documents retrieved for personalizing language model outputs.}
    \label{tab:supervised-time-dev}
\end{table*}

\begin{table*}
    \centering
    \scalebox{1}{
        \begin{tabular}{llcccc}
        \toprule
         & & \multicolumn{2}{c}{{Non-Personalized}} & \multicolumn{2}{c}{{Personalized}} \\
        \cmidrule(lr){3-4}\cmidrule(lr){5-6}
        {{Dataset}} & {{Metric}} & {FlanT5-XXL} & {GPT-3.5} & {FlanT5-XXL} & {GPT-3.5} \\\midrule
        
        \multirow{2}{*}{\shortstack[l]{{\benchmark-1T: Personalized}\\{Citation Identification}}} & \multirow{2}{*}{Accuracy $\uparrow$} & \multirow{2}{*}{0.498} & \multirow{2}{*}{0.478} & \multirow{2}{*}{\textbf{0.656}} & \multirow{2}{*}{0.640} \\\\\midrule

        \multirow{2}{*}{\shortstack[l]{{\benchmark-2T: Personalized}\\{Movie Tagging}}} & Accuracy $\uparrow$ & 0.326 & 0.333 & 0.361 & \textbf{0.439} \\
        & F1 $\uparrow$ & 0.255 & 0.273 & 0.283 & \textbf{0.403} \\\midrule

        \multirow{2}{*}{\shortstack[l]{{\benchmark-3T: Personalized}\\{Product Rating}}} & MAE $\downarrow$ & 0.335 & 0.720 & \textbf{0.294} & 0.608 \\
        & RMSE $\downarrow$ & 0.639 & 1.000 & \textbf{0.586} & 1.022 \\\midrule

        \multirow{2}{*}{\shortstack[l]{{\benchmark-4T: Personalized}\\{News Headline Generation}}} & ROUGE-1 $\uparrow$ & 0.173 & 0.146 & \textbf{0.192} & 0.159 \\
        & ROUGE-L $\uparrow$ & 0.157 & 0.128 & \textbf{0.175} & 0.138 \\\midrule

        \multirow{2}{*}{\shortstack[l]{{\benchmark-5T: Personalized}\\{Scholarly Title Generation}}} & ROUGE-1 $\uparrow$ & \textbf{0.472} & 0.413 & \textbf{0.472} & 0.421 \\
        & ROUGE-L $\uparrow$ & 0.419 & 0.348 & \textbf{0.422} & 0.352 \\\midrule

        \multirow{2}{*}{\shortstack[l]{{\benchmark-6T: Personalized}\\{Email Subject Generation}}} & ROUGE-1 $\uparrow$ & 0.316 & - & \textbf{0.382} & - \\
        & ROUGE-L $\uparrow$ & 0.302 & - & \textbf{0.369} & - \\\midrule

        \multirow{2}{*}{\shortstack[l]{{\benchmark-7T: Personalized}\\{Tweet Paraphrasing}}} & ROUGE-1 $\uparrow$ & \textbf{0.454} & 0.390 & 0.440 & 0.392 \\
        & ROUGE-L $\uparrow$ & \textbf{0.401} & 0.331 & 0.391 & 0.325 \\
        \bottomrule
        \end{tabular}
    }
    \caption{The zero-shot personalized text classification and generation results on the validation set of time-based separation setting. For personalized models, the tuned retriever based on the validation performance was selected.}
    \label{tab:llm-time-dev}
\end{table*}

\section{Performance of Some Other Non-Personalized Baselines on the \benchmark Benchmark}
\label{appendix:other-baselines}

To explore the performance of additional baselines, we present the performance of the Support Vector Machine (SVM) \cite{708428} as a conventional classifier, BERT \cite{Devlin2019BERTPO} as a neural transformer-based encoder, and BART \cite{bart} as a generative model. SVM and BERT are evaluated on classification tasks, and BART is evaluated on generation tasks within the \benchmark benchmark. The results for the user-based and time-based separation configurations are documented in Table \ref{tab:old-methods-user} and Table \ref{tab:old-methods-time}, respectively.

\begin{table*}
    \centering
    \scalebox{0.8}{
        \begin{tabular}{llcccccc}
        \toprule
         & & \multicolumn{2}{c}{{SVM}} & \multicolumn{2}{c}{{BERT}} & \multicolumn{2}{c}{{BART}} \\
        \cmidrule(lr){3-8}
        {{Dataset}} & {{Metric}} & {Validation} & {Test} & {Validation} & {Test} & {Validation} & {Test} \\\midrule
        
        \multirow{2}{*}{\shortstack[l]{{\benchmark-1T: Personalized}\\{Citation Identification}}} & \multirow{2}{*}{Accuracy $\uparrow$} & \multirow{2}{*}{0.512} & \multirow{2}{*}{0.523} & \multirow{2}{*}{0.520} & \multirow{2}{*}{0.483} & \multirow{2}{*}{-} & \multirow{2}{*}{-} \\\\\midrule

        \multirow{2}{*}{\shortstack[l]{{\benchmark-2T: Personalized}\\{Movie Tagging}}} & Accuracy $\uparrow$ & 0.836 & 0.609 & 0.609 & 0.593 & - & - \\
        & F1 $\uparrow$ & 0.810 & 0.580 & 0.588 & 0.559 & - & - \\\midrule

        \multirow{2}{*}{\shortstack[l]{{\benchmark-3T: Personalized}\\{Product Rating}}} & MAE $\downarrow$ & 0.227 & 0.515 & 0.395 & 0.364 & - & - \\
        & RMSE $\downarrow$ & 0.446 & 0.992 & 0.758 & 0.718 & - & - \\\midrule

        \multirow{2}{*}{\shortstack[l]{{\benchmark-4T: Personalized}\\{News Headline Generation}}} & ROUGE-1 $\uparrow$ & - & - & - & - & 0.166 & 0.159 \\
        & ROUGE-L $\uparrow$ & - & - & - & - & 0.151 & 0.145 \\\midrule

        \multirow{2}{*}{\shortstack[l]{{\benchmark-5T: Personalized}\\{Scholarly Title Generation}}} & ROUGE-1 $\uparrow$ & - & - & - & - & 0.418 & 0.409 \\
        & ROUGE-L $\uparrow$ & - & - & - & - & 0.380 & 0.375 \\\midrule

        \multirow{2}{*}{\shortstack[l]{{\benchmark-6T: Personalized}\\{Email Subject Generation}}} & ROUGE-1 $\uparrow$ & - & - & - & - & 0.405 & 0.379 \\
        & ROUGE-L $\uparrow$ & - & - & - & - & 0.387 & 0.361 \\\midrule

        \multirow{2}{*}{\shortstack[l]{{\benchmark-7T: Personalized}\\{Tweet Paraphrasing}}} & ROUGE-1 $\uparrow$ & - & - & - & - &  0.515 & 0.512 \\
        & ROUGE-L $\uparrow$ & - & - & - & - & 0.460 & 0.456 \\
        \bottomrule
        \end{tabular}
    }
    \caption{The results of non-personalized text classification and generation results on the validation and test set of user-based separation setting of SVM \cite{708428}, BERT \cite{Devlin2019BERTPO}, and BART \cite{bart}.}
    \label{tab:old-methods-user}
\end{table*}

\begin{table*}
    \centering
    \scalebox{0.8}{
        \begin{tabular}{llcccccc}
        \toprule
         & & \multicolumn{2}{c}{{SVM}} & \multicolumn{2}{c}{{BERT}} & \multicolumn{2}{c}{{BART}} \\
        \cmidrule(lr){3-8}
        {{Dataset}} & {{Metric}} & {Validation} & {Test} & {Validation} & {Test} & {Validation} & {Test} \\\midrule
        
        \multirow{2}{*}{\shortstack[l]{{\benchmark-1T: Personalized}\\{Citation Identification}}} & \multirow{2}{*}{Accuracy $\uparrow$} & \multirow{2}{*}{0.540} & \multirow{2}{*}{0.500} & \multirow{2}{*}{0.584} & \multirow{2}{*}{0.582} & \multirow{2}{*}{-} & \multirow{2}{*}{-} \\\\\midrule

        \multirow{2}{*}{\shortstack[l]{{\benchmark-2T: Personalized}\\{Movie Tagging}}} & Accuracy $\uparrow$ & 0.842 & 0.647 & 0.639 & 0.637 & - & - \\
        & F1 $\uparrow$ & 0.810 & 0.610 & 0.602 & 0.598 & - & - \\\midrule

        \multirow{2}{*}{\shortstack[l]{{\benchmark-3T: Personalized}\\{Product Rating}}} & MAE $\downarrow$ & 0.221 & 0.556 & 0.337 & 0.342 & - & - \\
        & RMSE $\downarrow$ & 0.448 & 1.203 & 0.706 & 0.725 & - & - \\\midrule

        \multirow{2}{*}{\shortstack[l]{{\benchmark-4T: Personalized}\\{News Headline Generation}}} & ROUGE-1 $\uparrow$ & - & - & - & - & 0.177 & 0.171 \\
        & ROUGE-L $\uparrow$ & - & - & - & - & 0.161 & 0.157 \\\midrule

        \multirow{2}{*}{\shortstack[l]{{\benchmark-5T: Personalized}\\{Scholarly Title Generation}}} & ROUGE-1 $\uparrow$ & - & - & - & - & 0.458 & 0.456 \\
        & ROUGE-L $\uparrow$ & - & - & - & - & 0.414 & 0.416 \\\midrule

        \multirow{2}{*}{\shortstack[l]{{\benchmark-6T: Personalized}\\{Email Subject Generation}}} & ROUGE-1 $\uparrow$ & - & - & - & - & 0.498 & 0.532 \\
        & ROUGE-L $\uparrow$ & - & - & - & - & 0.486 & 0.518 \\\midrule

        \multirow{2}{*}{\shortstack[l]{{\benchmark-7T: Personalized}\\{Tweet Paraphrasing}}} & ROUGE-1 $\uparrow$ & - & - & - & - &  0.507 & 0.504 \\
        & ROUGE-L $\uparrow$ & - & - & - & - & 0.455 & 0.453 \\
        \bottomrule
        \end{tabular}
    }
    \caption{The results of non-personalized text classification and generation results on the validation and test set of time-based separation setting of SVM \cite{708428}, BERT \cite{Devlin2019BERTPO}, and BART \cite{bart}.}
    \label{tab:old-methods-time}
\end{table*}

\section{Dataset Licenses}

This section specifies the licences and terms of use for each task, which is the same as the original dataset's license:

\begin{enumerate}[nolistsep, noitemsep]
    \item Personalized Citation Identification: CC BY-NC-SA 4.0
    \item Personalized Movie Tagging: Educational or academic research, NON COMMERCIAL USE
    \item Personalized Product Rating: CC BY-NC-SA 4.0
    \item Personalized News Headline Generation: CC BY-NC-SA 4.0
    \item Personalized Scholarly Title Generation: CC BY-NC-SA 4.0
    \item Personalized Email Subject Generation: Avocado Collection End User Agreement LDC2015T03
    \item Personalized Tweet Paraphrasing: CC BY-NC-SA 4.0
\end{enumerate}

\section{AI Assistance Usage}

In this paper, ChatGPT\footnote{\url{https://chat.openai.com/}} has been used as a writing assistant. In more detail, an initial paragraph is given to the ChatGPT and asked to paraphrase the given text. Further edits were also applied to the generated text and then used.

\end{document}